\documentclass{article} %
\usepackage[table]{xcolor}
\usepackage[final]{flageval_baai}
\usepackage[letterpaper,margin=1in]{geometry} %

\usepackage{amsmath,amsfonts,bm}

\def\eqref#1{equation~\ref{#1}}

\def\1{\bm{1}}

\DeclareMathAlphabet{\mathsfit}{\encodingdefault}{\sfdefault}{m}{sl}
\SetMathAlphabet{\mathsfit}{bold}{\encodingdefault}{\sfdefault}{bx}{n}

\usepackage{hyperref}
\usepackage{url}
\definecolor{darkblue}{rgb}{0, 0, 0.5}
\hypersetup{colorlinks=true, citecolor=darkblue, linkcolor=darkblue, urlcolor=darkblue}
\usepackage{graphicx}
\usepackage{booktabs} 
\usepackage{subcaption}
\usepackage{calc} 
\usepackage{caption}
\usepackage{amssymb}
\usepackage{adjustbox}
\usepackage{verbatim}
\usepackage{listings}
\usepackage[table]{xcolor}
\usepackage{subcaption}
\usepackage{wrapfig}
\usepackage{multirow}

\usepackage{tcolorbox}
\definecolor{lightgray}{HTML}{F0F0EB}
\definecolor{lightorange}{HTML}{FFD2A4}
\definecolor{lightblue}{HTML}{A4C7FF}
\definecolor{paleturquoise}{HTML}{AFEEEE}
\definecolor{lightgreen}{HTML}{A4FFAE}
\definecolor{stronggreen}{HTML}{3EFF54}
\definecolor{lightred}{HTML}{FFA4A4}
\definecolor{strongred}{HTML}{FF7171}

\lstdefinestyle{codeblock}{
  language=Python,
  basicstyle=\ttfamily\small,
  numbers=left,
  numberstyle=\tiny,
  stepnumber=1,
  numbersep=6pt,
  breaklines=true,
  frame=single,
  rulecolor=\color{black!20},
  backgroundcolor=\color{gray!5},
  tabsize=2,
  showstringspaces=false,
}

\fancyheadoffset[R,L]{0cm}

\title{Do Vision-Language Models Measure Up? Benchmarking Visual Measurement Reading with MeasureBench}

\author{
BAAI \textit{FlagEval} Team\thanks{Full list of authors attached at the end.} \\
\small \textbf{Project page:} \url{https://flageval-baai.github.io/MeasureBenchPage/}
}

\begin{document}

\maketitle

\begin{tcolorbox}[colback=paleturquoise, colframe=paleturquoise, 
  left=1pt,
  right=1pt,
  top=1pt,
  bottom=1pt]
{
\textbf{\emph{TL;DR:} Fine-grained visual understanding tasks such as visual measurement reading have been surprisingly challenging for frontier general-purpose vision-language models. We introduce MeasureBench, a benchmark with diverse images of measuring instruments collected from both real-world images and a new data synthesis pipeline.}
}
\end{tcolorbox}

\begin{abstract}
Reading measurement instruments is effortless for humans and requires relatively little domain expertise, yet it remains surprisingly challenging for current vision-language models (VLMs) as we find in preliminary evaluation. In this work, we introduce MeasureBench, a benchmark on visual measurement reading covering both real-world and synthesized images of various types of measurements, along with an extensible pipeline for data synthesis. Our pipeline procedurally generates a specified type of gauge with controllable visual appearance, enabling scalable variation in key details such as pointers, scales, fonts, lighting, and clutter. Evaluation on popular proprietary and open-weight VLMs shows that even the strongest frontier VLMs struggle with measurement reading in general.
We have also conducted preliminary experiments with reinforcement finetuning (RFT) over synthetic data, and find a significant improvement on both in-domain synthetic subset and real-world images. Our analysis highlights a fundamental limitation of current VLMs in fine-grained spatial grounding. We hope this resource and our code releases can help future advances on visually grounded numeracy and precise spatial perception of VLMs, bridging the gap between recognizing numbers and measuring the world.
\end{abstract}

\section{Introduction}
Recent advances in vision-language models (VLMs) have demonstrated impressive capabilities in tackling complex reasoning tasks that combine textual and visual information. Models or systems such as GPT-5 \citep{openai_gpt5_2025} and Gemini 2.5 Pro \citep{gemini25_2025} achieve human-expert level performance on college-level problems in MMMU \citep{yue2023mmmu} and MMMU-Pro \citep{yue2025mmmu-pro}. Even on Humanity's Last Exam (HLE) ~\citep{phan2025humanitysexam}, a benchmark measuring the frontier of human knowledge, state-of-the-art models achieve accuracies exceeding 25\%, substantially surpassing the human average.

That said, state-of-the-art VLMs still struggle with fine-grained perception, e.g., low-level visual cues, precise geometry, and subtle changes, even when their high-level reasoning appears strong.
Existing fine-grained evaluations are well represented by text reading and chart reasoning \citep{singh2019towards, masry-etal-2022-chartqa,tang2025chartmuseum}, or by similarly artificial low-level vision tests such as BlindTest \citep{vlms2024blind} and SalBench \citep{dahou2025salbench}. However, they rarely require mapping physical scales to numeric values.

Visual instrument reading tasks usually require fine-grained visual perception, light quantitative reasoning, and basic arithmetic operations.
Examples include reading pressure gauges in industrial settings, and thermometers or even as simple as clocks in daily life. Accurate interpretation of these instruments is crucial for safety, efficiency, and decision-making across domains for vision language models or future embodied AI systems. While a few existing studies have covered very specific types of reading such as clocks \citep{saxena2025lost,yang2022s}, rulers \citep{matuzevivcius2023rulers2023,pan2025reading}, industrial gauges \citep{izquierdo2025towards,valente2025cad2dmd}, and household meters \citep{van2025water}, they do not span the broad diversity of instruments or reading designs.

\begin{figure*}
    \centering
    \includegraphics[width=0.9\linewidth]{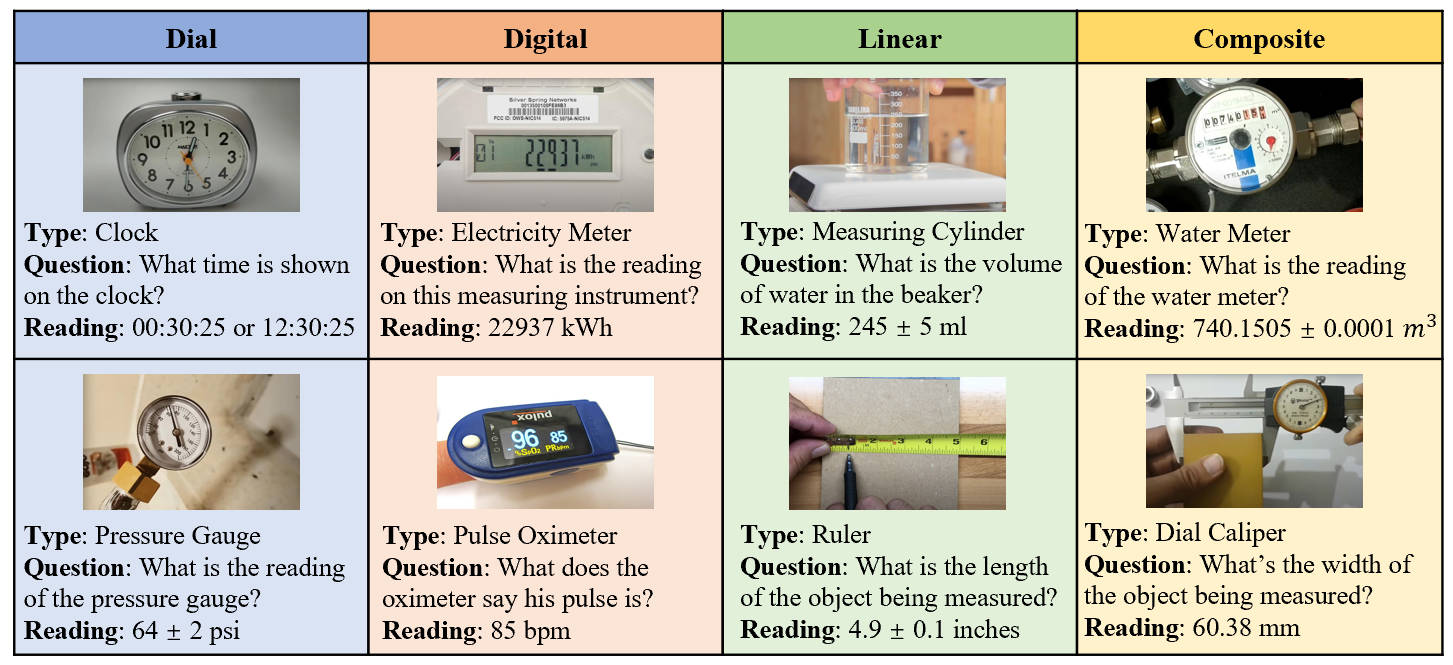}
    \caption{MeasureBench real-world samples with four commonly used reading designs.}
    \label{fig:intro-real}
\end{figure*}

To fill this gap, we introduce \textbf{MeasureBench}, a benchmark for evaluating VLMs on measuring instrument reading across 26 instrument types and four types of readout designs. Each image is paired with a reading question. MeasureBench comprises 2{,}442 image–question pairs: 1{,}272 diverse real-world images collected and human-annotated, and 1{,}170 synthetic images generated with randomized readings for 39 instruments.

Our data synthetic pipeline has two complementary backends: (i) a 2D programmatic renderer for diverse layouts with full control over fonts and geometry; and (ii) a 3D Blender renderer for photorealistic scenes with realistic lighting, materials, reflections, and occlusions. The pipeline is fully automated and readily scalable in both breadth (instrument types) and depth (variations). This pipeline can be used to generate additional data for training or evaluation.

We evaluate a number of modern VLMs on MeasureBench and report these key findings:
\begin{itemize}
\item \textbf{Persisting difficulty.} Current VLMs still struggle with instrument reading, with the best model achieving only 30.2\% accuracy on the real-world set and 26.3\% on the synthetic set.
\item \textbf{Object and text recognition seem easy, but inferring numbers is hard.} Models exhibit strong image understanding and reach over 90\% accuracy on unit recognition. Yet they falter on mapping scales to numeric values.
\item \textbf{Systematic fine-grained errors.} Models often ``know how to read'' but miss details: They misinterpret pointer positions, confuse adjacent ticks, and mismatch values to scale markings, leading to off-target answers.
\end{itemize}

With our data synthetic pipeline that produces accurately annotated readings, we have also conducted preliminary experiments of reinforcement learning using synthetic data.
Results are encouraging in that the synthetic subset of MeasureBench can get significantly improved, but not as promising on real-world images.

In summary, our contributions include:
\begin{itemize}
\item We present \textbf{MeasureBench}, a comprehensive benchmark targeting fine-grained instrument reading across 26 instrument types and 2,442 image–question pairs.
\item We provide a controllable 2D/3D synthesis pipeline that produces precise labels for sketch or photorealistic images with randomized readings for 39 instruments.
\item We deliver a \textbf{standardized evaluation} of 18 contemporary VLMs and an analysis of their failure modes, highlighting concrete gaps in low-level perception and precise geometric reasoning.
\item Preliminary RFT experiments using our synthetic pipeline show promise for data curation, but highlight the need for better visual representations to improve generalization.
\end{itemize}

\section{MeasureBench}

\subsection{Overview of MeasureBench}
We introduce \textit{MeasureBench}, a comprehensive benchmark for evaluating the ability to read values from measuring instruments. MeasureBench comprises two main components: (i) a diverse set of instrument images with standardized annotations, and (ii) a data synthesis framework for generating additional training and evaluation data.
By visual appearance, we categorize instruments into four readout designs (see also Figure~\ref{fig:intro-real} for examples from the real-world images in MeasureBench):

\begin{itemize}
    \item \textbf{Dial}: Analog instruments with one or more pointers (e.g., ammeters and pressure gauges which typically have a single pointer, whereas clocks often have two or three).
    \item \textbf{Digital}: Devices with electronic or mechanical digital readouts (e.g., pulse oximeters and electromechanical electricity meters).
    \item \textbf{Linear}: Instruments with linear scales and no pointers (e.g., rulers with a single scale, and vernier calipers with a main and a vernier scale).
    \item \textbf{Composite}: Instruments combining multiple readout designs, such as dial calipers and complex water meters.
\end{itemize}

As shown in Table~\ref{tab:measurebench-stats}, MeasureBench contains 2,442 questions: 1,272 real-world images and 1,170 synthetic images. The real-world subset spans 26 instrument types, while the synthetic subset covers 16 types with 39 distinct appearances.
To better explore the capability of VLMs in fine-grained instrument reading, we place greater emphasis on \emph{dial} and \emph{linear} instruments because digital devices primarily test OCR capabilities, and composite instruments are comparatively rare in practice.

\newlength{\panelh}
\setlength{\panelh}{5.2cm} 

\begin{figure}[t]
\centering
\setlength{\tabcolsep}{4pt}
\captionsetup{font=small,justification=centering,singlelinecheck=false,skip=2pt}

\begin{minipage}[t]{0.49\linewidth}
  \vbox to \panelh{%
    \centering\small
    \vfill
    \vfill
    \begin{tabular}{p{0.58\linewidth} r}
      \toprule
      \textbf{Statistics} & \textbf{Number} \\
      \midrule
      Total Questions & 2442 \\
      Real-World Images & 1272 (52\%) \\
      \textit{* Dial/Linear/Dig./Comp.} & 711/361/96/104 \\
      \textit{* Instrument Types} & 26 \\
      Synthetic Images & 1170 (48\%) \\
      \textit{* Dial/Linear/Dig./Comp.} & 750/300/60/60 \\
      \textit{* Instrument Types} & 16 \\
      \textit{* Instrument Appearances} & 39 \\
      \bottomrule
    \end{tabular}
    \vfill
  }
  \captionof{table}{Key statistics of MeasureBench.}
  \label{tab:measurebench-stats}
\end{minipage}\hfill
\begin{minipage}[t]{0.49\linewidth}
  \vbox to \panelh{
    \centering
    \includegraphics[height=\panelh,keepaspectratio]{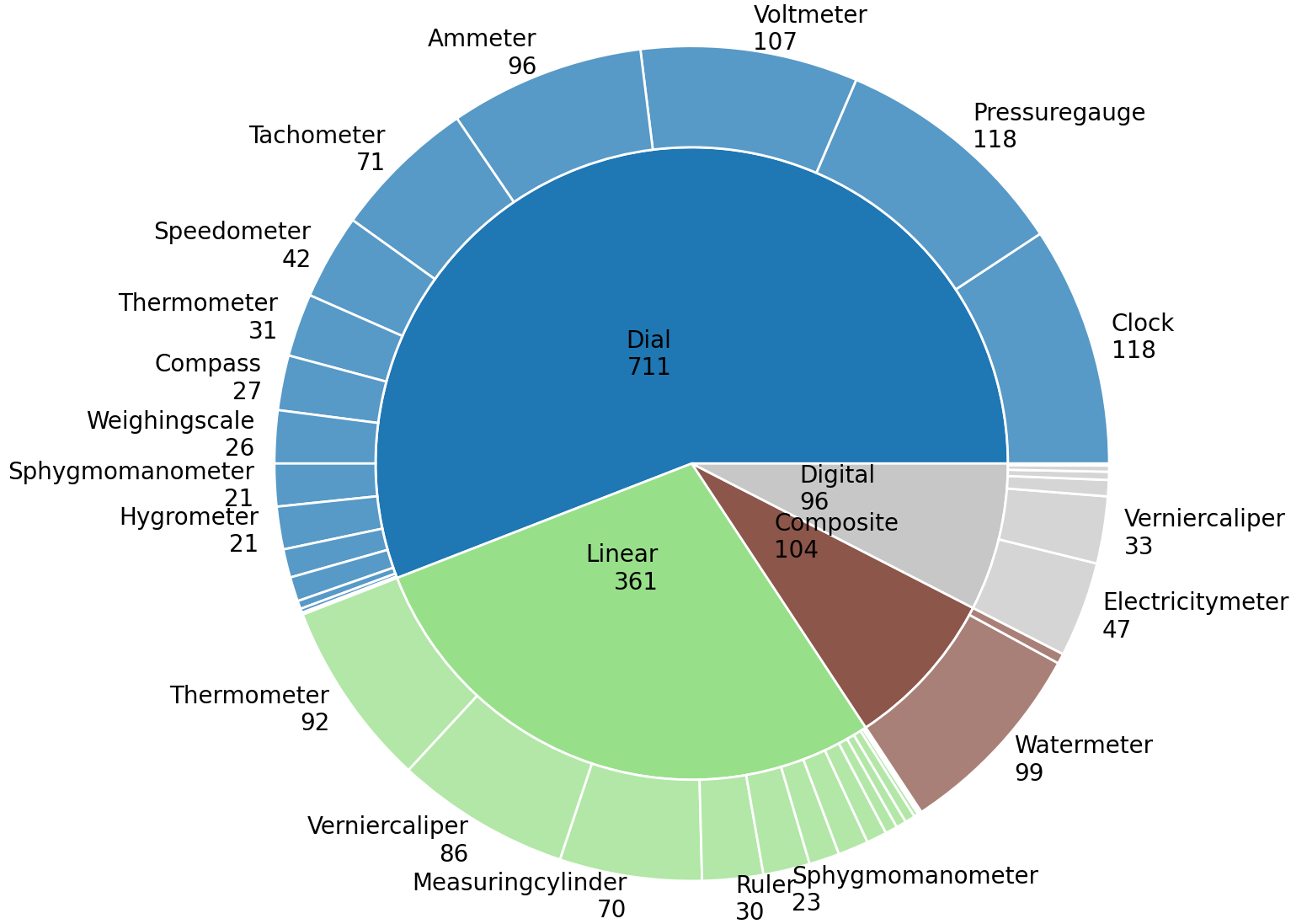}
    \vfill
  }
  \captionof{figure}{Distribution of reading designs and instrument types.}
  \label{fig:design_analysis}
\end{minipage}
\end{figure}

\subsection{Evaluation metrics}
\label{sec:eval_metrics}
Measurement error is natural when reading from any instrument that does not explicitly display a deterministic digital value on the screen.
Therefore, we determine the correctness of the final reading via interval matching instead of a strict value, along with the correctness of unit prediction.

\textbf{Answer extraction.} To get the reading from natural language output, we extract the final answer after common markers (e.g., “Answer:”) or inside \verb|\boxed{}|. Our evaluation script will specifically parse:
(i) \emph{numeric}: integers, decimals, scientific notation, and fractions (\verb|a/b|$\to$float). If multiple scalars appear, use the rightmost.
(ii) \emph{time}: the first \texttt{hh:mm[:ss]} pattern, converted to seconds.
Preserve adjacent tokens for unit matching.
\footnote{Unicode characters are normalized for equivalence matching.}

\textbf{Answer matching.} Each sample in our benchmark includes one or more ground-truth candidates\footnote{Some instruments express the same quantity in different units thereby different values, such as degrees in Celsius and Fahrenheit. During evaluation, we only adopt the single ground-truth that maximizes the score.},
each contains a closed numeric interval for value grading, optionally along with a set of acceptable unit substrings to indicate a correct unit in a model response.
A prediction is \emph{value-correct} if the number parsed from the model response \emph{falls within the interval} of any candidate, and \emph{unit-correct} if any unit string in that candidate can match the extracted answer. \emph{Fully-correct} requires value-correct and, when specified, unit-correct for the \emph{same} candidate.
If multiple candidates exist, score against the one that maximizes correctness (prefer fully-correct; otherwise prefer value-correct; break ties by smaller relative error, then by narrower interval).

\subsection{Real-world subset curation}
We assembled a real-world subset of images from three sources: (i) Google Image Search using instrument-specific keywords, restricted to images under permissive licenses for usage, (ii) photos contributed by team members under private authorization, and (iii) images purchased from a third-party vendor. We removed low-quality images (e.g., blurry, low-resolution, or occluded) and annotated the remaining images using a standardized schema. For each image, we recorded the instrument type, readout design, candidate units, and the valid interval of reading values; any value within this interval is considered correct.

We recruited 10 qualified annotators and assigned tasks aligned with their professional backgrounds. Each image was independently labeled by one annotator and verified by another; disagreements were adjudicated by a third annotator.
Another independent round of review was conducted to verify the correctness of annotation, including the numerical intervals and the unit.
We have also conducted a preliminary analysis on prompt sensitivity (with details in Sec~\ref{sec:prompt-sensitivity}) and find very little impact on overall results, so we leave most of the collected prompts unchanged.

\subsection{Data synthesis framework}
\begin{figure*}[!ht]
    \centering
    \includegraphics[width=0.9\linewidth]{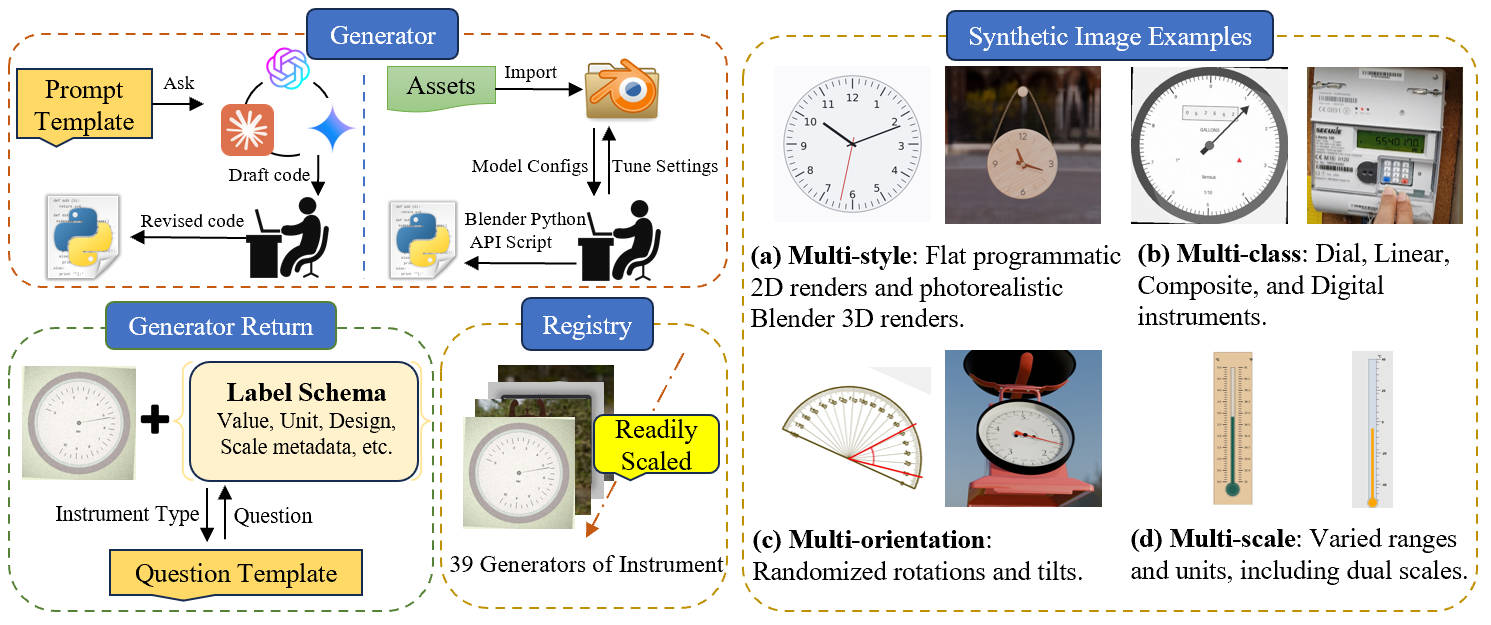}
    \caption{Left: A hybrid measuring instrument synthesis framework. Right: Examples of synthetic images.}
    \label{fig:synthetic_pipeline}
\end{figure*}

As shown in Figure \ref{fig:synthetic_pipeline}, we develop a modular and scalable synthesis framework that treats each instrument as a \textbf{generator} registered under a unified interface.  
A global \textbf{registry} maps instrument names to generators; each generator returns a rendered image together with a standardized label schema comprising the numeric \emph{value}, \emph{unit} and \emph{readout design}.  
This uniform contract enables plug-and-play additions.

For each sample, the framework randomizes
the number and type of scales, measurement readout numbers, ranges/units,  materials, lighting, backgrounds, and camera pose, while enforcing semantic validity.  
Figure~\ref{fig:synthetic_pipeline} illustrates the resulting diversity along four axes.

We provide two complementary back-ends under the same interface:  
\begin{itemize}
    \item \textbf{2D programmatic rendering.} 
    A prompt template specifies instrument types and reading constraints, and prescribes the code interface and preferred libraries. LLMs then draft the code accordingly. We verify the code before registering the generator.%
    \item \textbf{3D physical rendering.} We leverage existing Blender\footnote{https://www.blender.org/} assets and write code to randomize backgrounds, instrument readings (e.g., pointer angles, scale ranges) and camera pose to produce photorealistic images and narrow the sim-to-real gap. (See appendix for details)
\end{itemize}

We implement \textbf{39} distinct appearances spanning \textbf{16} instrument types. For benchmarking purpose, we independently generate \textbf{30} images per appearance, totaling \textbf{1,170} synthetic images. As illustrated in Figure \ref{fig:synthetic_pipeline}, the images vary along four axes:
\emph{multi-style} (2D vs.\ photorealistic 3D), 
\emph{multi-class} (dial, linear, composite, digital),
\emph{multi-orientation} (rotations/tilts and imaging perturbations), and 
\emph{multi-scale} (ranges/units and dual scales), providing broad coverage for robust reading models.

\section{Evaluation Results}
We present a systematic evaluation of various VLMs on MeasureBench: 8 proprietary and 10 open-weight models. The evaluated model families include GPT~\citep{openai_gpt5_2025}, Claude~\citep{claude4_system_card}, Gemini~\citep{gemini25_2025}, Mistral~\citep{mistral_medium_3}, Grok~\citep{grok_4_xai}, Qwen-VL~\citep{Qwen2.5-VL}, InternVL3~\citep{zhu2025internvl3}, and LLaMA-4 \citep{llama4_models_page}.
All evaluations are conducted using FlagEvalMM~\citep{he-etal-2025-flagevalmm}, a flexible and comprehensive framework for multimodal model evaluation.

\subsection{Main results}

\begin{table*}[t]
\centering
\small
\rowcolors{2}{gray!10}{white}
\begin{tabular}{l|ccccccc|ccccccc}
\toprule
\textbf{Model} & \multicolumn{7}{c|}{\textbf{Real-world subset}} & \multicolumn{7}{c}{\textbf{Synthetic subset}} \\
 & \textbf{Ovr} & \textbf{Val} & \textbf{Unit} & \textbf{Dial} & \textbf{Dig} & \textbf{Lin} & \textbf{Com} & \textbf{Ovr} & \textbf{Val} & \textbf{Unit} & \textbf{Dial} & \textbf{Dig} & \textbf{Lin} & \textbf{Com} \\
\midrule
Gemini-2.5-Pro & \textbf{30.2} & \textbf{30.7} & \textbf{96.2} & \textbf{31.5} & \textbf{80.2} & \textbf{21.9} & \textbf{3.8} & \textbf{26.3} & \textbf{26.8} & 93.1 & \textbf{18.3} & 70.0 & \textbf{40.0} & \textbf{15.0} \\
Qwen3-VL-235B & 22.6 & 23.0 & 95.7 & 23.5 & 64.6 & 15.2 & 2.9 & 19.0 & 19.6 & 94.4 & 14.1 & 60.0 & 26.3 & 1.7 \\
GPT-5-Mini & 22.0 & 22.4 & 95.2 & 20.8 & 70.8 & 16.9 & 2.9 & 17.9 & 18.6 & 93.2 & 12.0 & 56.7 & 28.3 & 1.7 \\
Gemini-2.5-Flash & 20.2 & 21.1 & 93.4 & 20.5 & 65.6 & 13.0 & 1.0 & 18.1 & 19.0 & 91.7 & 11.9 & \textbf{75.0} & 25.7 & 1.7 \\
GPT-5 & 19.8 & 19.9 & 96.0 & 18.3 & 66.7 & 15.2 & 2.9 & 16.9 & 17.5 & 94.3 & 9.7 & 48.3 & 31.7 & 1.7 \\
Qwen3-VL-8B & 15.3 & 15.8 & 94.0 & 14.5 & 53.1 & 11.3 & 0.0 & 11.4 & 11.6 & 92.4 & 8.0 & 25.0 & 19.3 & 0.0 \\
Qwen2.5-VL-7B & 14.6 & 15.0 & 93.4 & 13.8 & 49.0 & 11.4 & 0.0 & 10.9 & 11.5 & 88.5 & 5.7 & 33.3 & 21.7 & 0.0 \\
Qwen2.5-VL-72B & 14.5 & 14.9 & 92.1 & 12.2 & 55.2 & 12.2 & 0.0 & 11.7 & 12.0 & 92.3 & 6.4 & 43.3 & 21.0 & 0.0 \\
Claude-Opus-4.1 & 14.3 & 14.9 & 94.5 & 14.8 & 38.5 & 11.1 & 0.0 & 13.3 & 14.1 & 93.1 & 6.4 & 45.0 & 27.0 & 0.0 \\
InternVL3.5-38B & 12.9 & 13.6 & 89.8 & 12.1 & 51.6 & 7.7 & 0.0 & 12.6 & 15.4 & 78.5 & 6.3 & 41.7 & 25.3 & 0.0 \\
Claude-Sonnet-4 & 12.6 & 13.1 & 89.9 & 15.0 & 20.8 & 9.1 & 0.0 & 11.0 & 11.5 & 92.8 & 5.1 & 26.7 & 25.0 & 0.0 \\
LLaMA-4-maverick & 12.2 & 12.9 & 91.6 & 12.1 & 44.8 & 7.2 & 0.0 & 12.1 & 13.2 & 89.7 & 6.3 & 50.0 & 21.7 & 0.0 \\
Qwen2.5-VL-32B & 11.7 & 12.0 & 94.6 & 9.0 & 51.6 & 9.7 & 0.0 & 10.5 & 10.7 & \textbf{96.0} & 5.3 & 28.3 & 22.0 & 0.0 \\
LLaMA-4-scout & 10.9 & 11.4 & 90.6 & 8.2 & 54.2 & 8.0 & 0.0 & 9.1 & 10.2 & 86.4 & 5.5 & 20.0 & 17.7 & 0.0 \\
Mistral-medium-3.1 & 10.6 & 11.2 & 93.4 & 7.0 & 57.3 & 8.3 & 0.0 & 8.5 & 8.8 & 91.6 & 3.7 & 23.3 & 19.3 & 0.0 \\
InternVL3.5-8B & 9.7 & 10.9 & 84.0 & 10.4 & 30.5 & 5.5 & 0.0 & 7.7 & 8.4 & 84.6 & 3.5 & 26.7 & 16.0 & 0.0 \\
Mistral-small-3.2 & 8.5 & 9.7 & 81.3 & 7.9 & 32.3 & 5.8 & 0.0 & 6.5 & 8.0 & 80.5 & 3.2 & 5.0 & 16.3 & 0.0 \\
Grok-4 & 7.5 & 7.7 & 80.5 & 6.5 & 24.0 & 7.5 & 0.0 & 6.2 & 6.4 & 71.6 & 3.3 & 25.0 & 10.3 & 1.7 \\
\bottomrule
\end{tabular}
\caption{Performance on real and synthetic images. We report accuracy (\%) for each model: overall (Ovr), value (Val), unit (Unit), and by readout type—Dial, Digital (Dig), Linear (Lin), Composite (Com).}
\label{tab:combined_measurebench}
\end{table*}

Table \ref{tab:combined_measurebench} reports results on MeasureBench.
The best-performing \emph{Gemini 2.5 Pro} reaches only 30.2\% overall accuracy on real images and 26.3\% on synthetic images, showing that reading measuring instruments remains a challenging fine-grained vision task for current VLMs.
A few more observations:

\textbf{\emph{Value} reading is the bottleneck.}
Across models, recognizing (or inferring) the unit is consistently above 90\% accurate on both real and synthetic sets, while value accuracy is much lower (e.g., Gemini 2.5 Pro: \textbf{96.2}\% unit vs.\ \textbf{30.7}\% value on real images). This suggests strong OCR or object recognition capabilities from current VLMs for unit prediction. However, the low performance shows that estimating the numerical \emph{value} often requires precise localization of pointers, ticks, and scales.

\textbf{Different readout designs are not equally challenging.}
Table \ref{tab:combined_measurebench} displays decomposed accuracy metrics by instrument type. \emph{Digital} displays are much easier (e.g., up to 80.2\% on real images for Gemini 2.5 Pro), reflecting reliance on OCR. \emph{Dial} and \emph{linear} instruments remain challenging (typically 10-32\%), as they require needle localization or reading tick marks under clutter, highlights, and distortion. \emph{Composite} instruments are by far the most challenging: they require combining multiple readout designs, reading each component correctly, and performing the corresponding numerical calculations.

\textbf{Larger models may not always read better.}
Table \ref{tab:combined_measurebench} shows that GPT-5-Mini marginally outperforms GPT-5, while Qwen2.5-VL-7B performs on par with Qwen2.5-VL-72B and outperforms Qwen2.5-VL-32B. Through case studies (more in Appendix), we observe that while a portion of GPT-5-Mini's correct answers can attribute to successful guessing, GPT-5 genuinely erred in image recognition on a different subset of problems.
Within the Qwen family, overall performance does not monotonically improve with larger LLMs when the visual encoder remains unchanged.\footnote{Qwen2.5-VL-7B, Qwen2.5-VL-32B and Qwen2.5-VL-72B use a ViT with the same number of parameters~\citep{Qwen2.5-VL}.} This implies that larger language backbones do not contribute to better fine-grained perception,
while sometimes the language prior may incorrectly bias the predicted values.
For instance, Qwen2.5-VL-72B has shown a much stronger preference in predicting ``10:10'' on clock images (73\% of real images), as described in Appendix~\ref{app:1010}.

\textbf{Real vs.\ synthetic.}
The gap between real and synthetic is modest but consistent: most VLMs achieve generally lower accuracy on synthetic images than real images (e.g., Gemini-2.5-Pro 30.2 to 26.3, GPT-5 19.8 to 16.9), showing that synthetic scenes remain a genuine challenge rather than an easier abstraction. The drop is mainly driven by \emph{value} accuracy, while \emph{unit} accuracies remain similar, suggesting that numeric extraction is the primary failure mode. %

\begin{figure}[!ht]
    \centering
    \includegraphics[width=\linewidth]{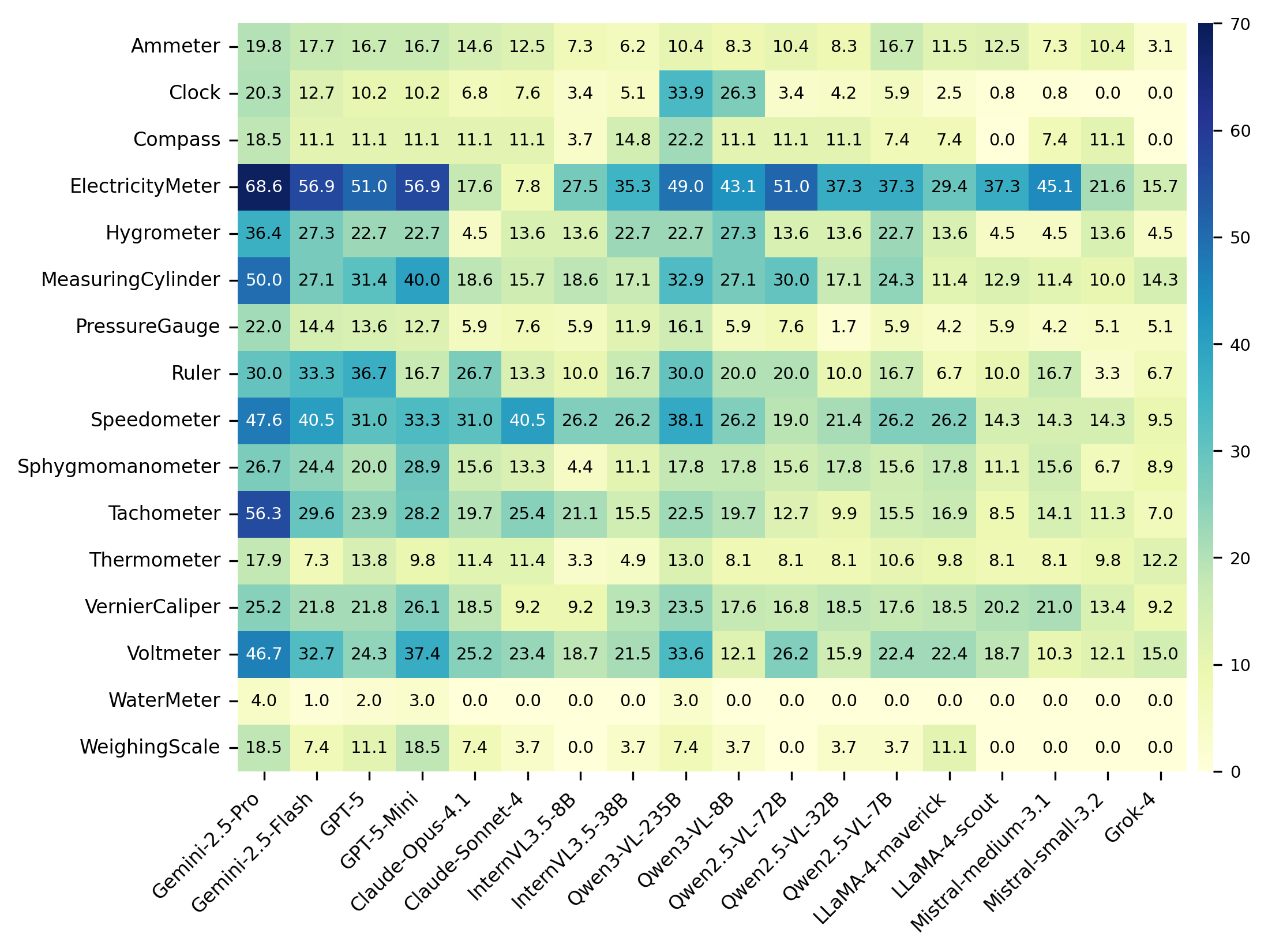}
    \caption{Model accuracies on real images by instrument category (categories with $\geq$20 samples).}
    \label{fig:heatmap}
\end{figure}
\textbf{Category-wise performance varies widely.}
Figure \ref{fig:heatmap} reveals substantial spread across instrument categories. Categories with a high proportion of digital readouts (e.g., electricity meters) tend to achieve higher accuracy, whereas categories dominated by multi-needle dials (e.g., clocks, water meters) are challenging for all models. Single-needle dials with sparse ticks (e.g., speedometer) are generally easier, and linear gauges (e.g., rulers, measuring cylinders) are easier overall than dials.

\subsection{To think, or not to think?}

\begin{wrapfigure}{R}{0.6\textwidth}
    \vspace{-6mm}
    \centering
    \includegraphics[width=7cm]{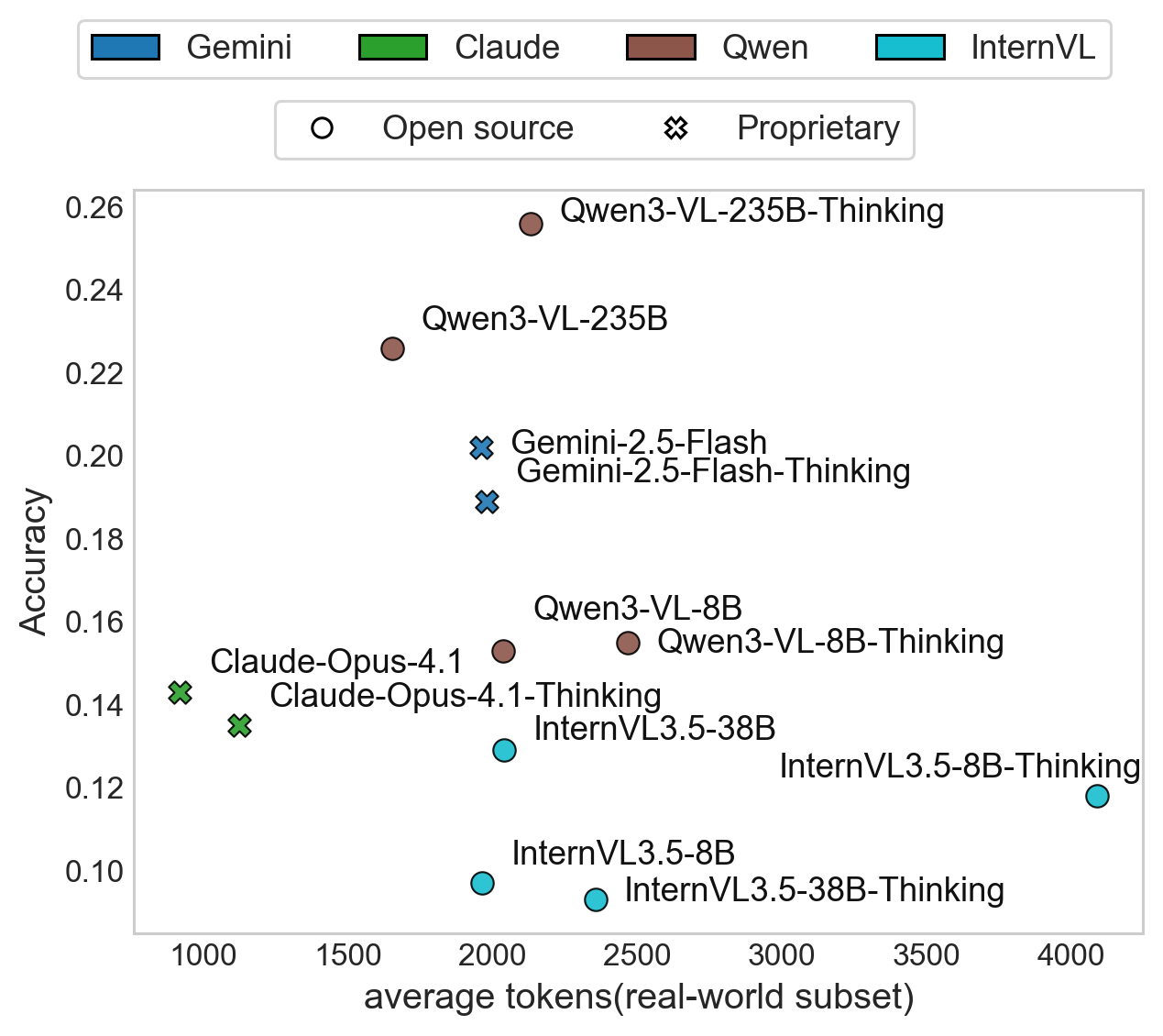}
    \vspace{-4mm}
    \caption{Performance and efficiency analysis of various large vision-language models. The accuracy against the average token count is plotted to show the performance-cost trade-off.}
    \label{fig:scatter}
    \vspace{-4mm}
\end{wrapfigure}

Inference-time ``thinking'' has been widely adopted to improve large language models (LLMs) on complex text-based reasoning. We ask whether this also holds for VLMs on \emph{MeasureBench}, which demands fine-grained visual perception coupled with numerical reasoning. We compare a couple of \emph{hybrid reasoning} models under a \emph{no-thinking} setting (reasoning tokens set to $0$) against a \emph{thinking} setting (maximum of $10,240$ reasoning tokens). The study covers six models: InternVL3.5-8B/38B, Qwen3-VL-8B/235B, Claude 4.1 Opus, and Gemini 2.5 Flash.

As shown in Figure~\ref{fig:scatter}, enabling thinking yields very little improvement, sometimes even degrades performance. While thinking often boosts text-only reasoning, it does not appear to help VLMs attend to the most relevant image regions or to enhance fine-grained visual perception on MeasureBench.
This conforms to recent findings on the utility of test-time thinking for visual problems~\citep{FlagEval2025LRM-Eval}.

Figure~\ref{fig:scatter} further relates accuracy to the average number of reasoning tokens consumed per sample. Although thinking increases token usage, the increment is modest (a few hundreds up to 1k-2k tokens), yet accuracy gains remain limited. Instrument reading primarily requires precise visual interpretation rather than extended chain-of-thought, so extended text-based reasoning is ineffective to improve performance on this task.

\subsection{Prompt sensitivity analysis}
\label{sec:prompt-sensitivity}
We investigate whether prompting style affects instrument reading by comparing two styles:

\begin{itemize}
    \item \textbf{Specific}: explicitly names the instrument type (e.g., \emph{“What is the reading of this ammeter?”}).
    \item \textbf{General}: uses a uniform query (e.g., \emph{“What is the reading of the instrument?”}), requiring the model to infer the instrument type and the quantity being measured.
\end{itemize}

A manual review of prompts yields three categories:
besides \emph{specific} (9.36\%) and \emph{general} (80.11\%), the remaining prompts are \emph{immutable} (10.53\%) as essential information prevents rewriting (e.g., ``What is the high pressure according to the sphygmomanometer?''). 
We exclude them and rewrite the other two sets into the two styles.
We also explore the effect of adding a prompt suffix: ``\textit{Provide an answer as precise as possible}''.

As shown in Table~\ref{tab:prompt-sensitivity},
prompting with specific instrument names only has a moderately positive impact on unit accuracy but no real difference on total accuracy. Meanwhile, explicitly asking VLMs to provide as precise answers does not yield consistent impact on performance either.
Hence we retain the original simple prompts for main benchmark evaluation without additional prompt engineering.

\begin{table}[!t]
\centering
\small
\rowcolors{2}{gray!10}{white}
\begin{tabular}{lccccc}
\toprule
\textbf{Model} & \textbf{Precise} & \multicolumn{2}{c}{\textbf{Total Acc.}} & \multicolumn{2}{c}{\textbf{Unit Acc.}} \\
\cmidrule(lr){3-4} \cmidrule(lr){5-6}
 & & \textbf{Gen.} & \textbf{Spec.} & \textbf{Gen.} & \textbf{Spec.}\\
\midrule
GPT-4.1-Mini & No & 15.2 & 17.7 & 88.0 & 90.0 \\
GPT-4.1-Mini & Yes & 15.2 & 15.5 & 88.8 & 91.4 \\
InternVL3.5-8B & No & 9.8 & 9.6 & 84.5 & 90.2 \\
InternVL3.5-8B & Yes & 11.6 & 10.8 & 84.2 & 89.6 \\
Qwen2.5-VL-72B & No & 13.5 & 14.9 & 90.1 & 90.7 \\
Qwen2.5-VL-72B & Yes & 14.0 & 13.4 & 91.5 & 92.8 \\
\bottomrule
\end{tabular}
\caption{Model performance under different prompt settings.}
\label{tab:prompt-sensitivity}
\end{table}

\subsection{Case studies}

\begin{figure*}[!ht]
    \centering
    \includegraphics[width=1.0\linewidth]{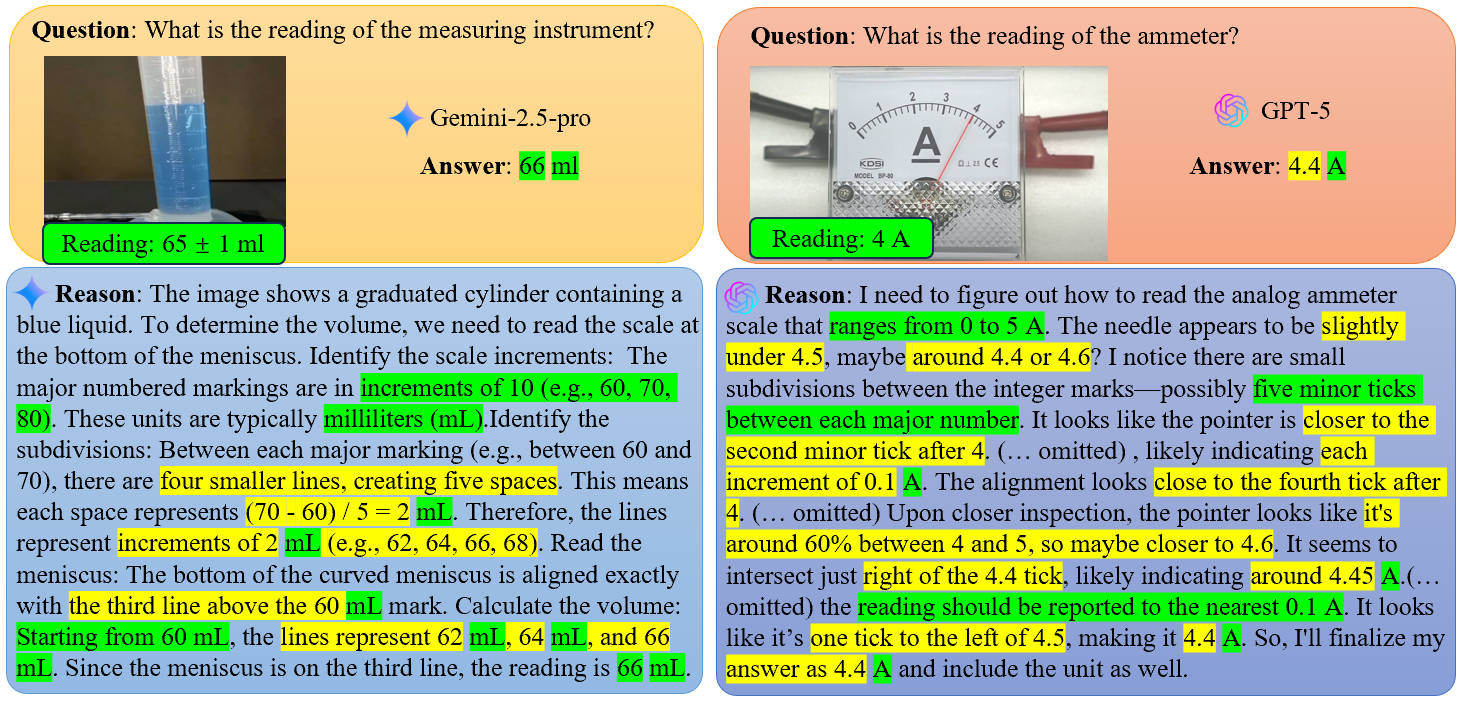}
    \caption{Case studies. \colorbox{green}{Text in green} marks statements consistent with the image; \colorbox{yellow}{yellow} marks contradictions.}
    \label{fig:bad_case}
\end{figure*}

Figure~\ref{fig:bad_case} shows two typical examples from our benchmark: a measuring cylinder and an ammeter. In each panel, \colorbox{green}{text in green} highlights denote statements consistent with the image, whereas \colorbox{yellow}{text in yellow} denotes claims that are contradicted by the visual evidence.

\textbf{What VLMs get right:}
Models generally \emph{know the task}. They identify the instrument, locate the indicator (meniscus/needle), infer the unit and major tick spacing, and try to interpolate to a final value. This shows mission awareness and an active search for the pointer.

\textbf{Where they fail:}
Most errors arise from small perceptual mistakes that dominate the numeric outcome:
(i) \emph{Pointer localization}: one minor tick left/right changes the reading (e.g., 4.4 vs.\ 4.5~A).
(ii) \emph{Indicator interpretation}: wrong minor-tick count or reading the wrong edge of the meniscus.

\textbf{Right answer, wrong reasoning.}
We observe frequent \emph{error cancellation}. In the cylinder example (Gemini-2.5-pro), an incorrect subdivision story coincidentally offsets a later mistake, yielding the correct number. Such cases inflate accuracy if only the final answer is scored.

\subsection{Impact of detailed instructional prompts}
We investigate whether providing step-by-step reading instructions tailored to different instrument types can improve model performance.
The instruction includes design-level common rules shared by instruments of the same readout design and instrument-specific guidance.
As shown in Table~\ref{tab:detailed_prompt_comparison}, even with explicit guidance on how to read the instruments, the performance gain on real-world MeasureBench is very limited. Adding in-context examples does not seem to help either, suggesting that the bottleneck lies in fine-grained visual perception rather than a lack of procedural knowledge.

\begin{table}[!ht]
\centering
\small
\rowcolors{2}{gray!10}{white}
\begin{tabular}{l|ccc}
\toprule
\textbf{Model} & \textbf{Overall} & \textbf{Value} & \textbf{Unit}\\
\midrule
Gemini-2.5-Pro & 31.8(+1.6) & 32.2(+1.5) & 96.8(+0.6) \\
GPT-5 & 19.9(+0.1) & 20.1(+0.2) & 97.5(+1.5) \\
Qwen2.5-VL-7B & 14.6(+0.0) & 15.5(+0.5) & 93.2(-0.2) \\
\bottomrule
\end{tabular}
\caption{Impact of more detailed instructional prompts on MeasureBench (real-world images). Numbers in parentheses indicate the change compared to the default prompt.}
\label{tab:detailed_prompt_comparison}
\end{table}

\subsection{Generalization on general benchmarks}
To assess potential negative transfer from reinforcement finetuning on synthetic measurement data, we evaluate Qwen2.5-VL-7B before and after GRPO training on popular general-purpose benchmarks. As shown in Table~\ref{tab:qwen_comparison}, GRPO training on synthetic data yields comparable performance on these benchmarks with no degradation, indicating that the model retains its general capabilities.

\begin{table}[!ht]
\centering
\small
\resizebox{0.48\textwidth}{!}{%
\begin{tabular}{l|cccc}
\toprule
\textbf{Model} & \textbf{MMMU} & \textbf{MMMU-Pro} & \textbf{MathVista} & \textbf{TextVQA} \\
\midrule
Qwen2.5-VL-7B & 52.44 & 37.75 & \textbf{69.40} & \textbf{84.74} \\
+ GRPO & \textbf{54.33} & \textbf{37.86} & 69.00 & 84.24 \\
\bottomrule
\end{tabular}
}
\caption{Qwen2.5-VL-7B results on general benchmarks before and after GRPO training on synthetic measurement data.}
\label{tab:qwen_comparison}
\end{table}

\subsection{Comparison with supervised fine-tuning}
We compare GRPO with supervised fine-tuning (SFT) on the same synthetic dataset. We experiment with two SFT response formats: (i) direct answer only (e.g., ``4.3\,cm'', ``6.5\,L''), and (ii) rationale + final answer, where GPT-5.2 is used to generate step-by-step rationales given the image and the ground-truth readout. As shown in Table~\ref{tab:sft_grpo_comparison}, while SFT substantially improves in-domain synthetic performance, it degrades accuracy on real-world images, indicating severe overfitting to synthetic patterns. In contrast, GRPO improves on both synthetic and real-world subsets, demonstrating better generalization.

\begin{table}[!ht]
\centering
\small
\rowcolors{2}{gray!10}{white}
\begin{tabular}{l|ccc|ccc}
\toprule
\multirow{2}{*}{\textbf{Model}}
& \multicolumn{3}{c|}{\textbf{Real-world}}
& \multicolumn{3}{c}{\textbf{Synthetic}}\\
\cmidrule(lr){2-4}\cmidrule(lr){5-7}
& \textbf{Ovr} & \textbf{Val} & \textbf{Unit}
& \textbf{Ovr} & \textbf{Val} & \textbf{Unit}\\
\midrule
Qwen2.5-VL-7B & 14.6 & 15.0 & \textbf{93.4} & 10.9 & 11.5 & 88.5 \\
+ GRPO & \textbf{19.7} & \textbf{20.4} & 92.3 & \textbf{35.2} & \textbf{35.6} & 96.7 \\
+ answer SFT & 12.5 & 13.4 & 89.5 & 29.7 & 30.9 & \textbf{96.8} \\
+ rationale SFT & 8.8 & 9.1 & 91.3 & 21.7 & 22.0 & 95.8 \\
\bottomrule
\end{tabular}
\caption{Comparison of GRPO and SFT training on Qwen2.5-VL-7B. SFT overfits to synthetic patterns, degrading real-world performance, while GRPO improves both.}
\label{tab:sft_grpo_comparison}
\end{table}

\subsection{Would earlier special-purpose systems work?}
Prior to the emergence of general-purpose VLMs, there exist
earlier domain-specific computer vision systems~\citep{shu2023read,reitsma2024under} designed for gauge reading where model checkpoints are still available. These systems manually design specialized pipelines, e.g., detecting the gauge region, localizing the pointer, recognizing scale marks, and reading text via OCR.
We have verified that the systems are performing as expected on their test sets, before evaluating both systems on a relevant subset (dial meters) of MeasureBench. The results show very little generalization across detection, pointer localization, or text reading on our benchmark data which may differ a lot from their training images. As pipeline systems, they tend to fail on the new images at different components including gauge detection, number recognition (OCR), and pointer segmentation.

Table \ref{tab:cvReading} shows that the generalization of pointer value detection is inferior to that of general-purpose VLMs. The special-purpose neural network components \citep{shu2023read} more or less overfit their training data, resulting in failures to detect pointers or scale marks on most out-of-distribution datasets. 
Meanwhile, the accuracy of OCR-based unit recognition \citep{reitsma2024under} is significantly lower than that of VLMs.

\begin{table}[!ht]
\centering
\small
\rowcolors{2}{gray!10}{white}
\begin{tabular}{l|ccc|ccc}
\toprule
\multirow{2}{*}{\textbf{Model}}
& \multicolumn{3}{c|}{\textbf{Real-world }}
& \multicolumn{3}{c}{\textbf{Synthetic }}\\
\cmidrule(lr){2-4}\cmidrule(lr){5-7}
& \textbf{Ovr} & \textbf{Val} & \textbf{Unit}
& \textbf{Ovr} & \textbf{Val} & \textbf{Unit}\\
\midrule
Reitsma et al. \cite{reitsma2024under} & 8.5 & 11.9 & 17.8 & 12.5 & 12.5 & 17.5 \\
Shu et al. \cite{shu2023read}     & N/A & 4.2  & N/A  & N/A  & 0.0  & N/A  \\
Gemini-2.5-Pro         & 30.2 & 30.7 & 96.2 & 26.3 & 26.8 & 93.1 \\
Qwen3-VL-8B          & 15.3  & 15.8 & 94.0 & 11.4 & 11.6 & 92.4 \\
\bottomrule
\end{tabular}
\caption{Model accuracy (\%) for VLMs and prior special-purpose systems; ``N/A'' = failed on all examples.}
\label{tab:cvReading}
\end{table}

\section{Can training with synthetic data help?}
\label{sec:train_grpo}
The data synthesis pipeline provides diverse instrument images, which raises the question of whether training on synthetic data can improve performance on real-world images. To investigate this, we generated 100 samples for each of the 39 instruments (3,900 image–question pairs) and used them for model training.
The task format especially suits reinforcement learning via assigning a positive reward on correct reading results.
We adapt the GRPO algorithm~\citep{shao2024deepseekmath} to conduct reinforcement finetuning (RFT) on \textsc{Qwen2.5-VL-3B} and \textsc{Qwen2.5-VL-7B}. Models are trained using the verl library \cite{sheng2024hybridflow}, training details are listed in appendix.

\subsection{Reward design}
To stay consistent with the scoring used in our evaluation, we employ a rule-based reward aligned with the evaluation method.\footnote{We have also tried a soft variant that gives partial reward to predicted values that are close to the ground-truth interval, but not observing much difference. We discuss more details in the Appendix.}
Let $I=[l,r]$ and $u$ denote the ground-truth interval and unit, and let $\tilde{p}$ be the model's textual prediction from which we extract the numeric value $\hat{y}$ and unit $\hat{u}$. $\mathcal{F}$ is the response pattern ``\texttt{<think>.*</think>.*Final Answer.*}''
Define the indicators
\begin{align}
c_{\mathrm{all}} &= \mathbb{I}\{\hat{y}\in I \land \hat{u}=u\} \label{eq:call}\\
c_{\mathrm{fmt}} &= \mathbb{I}\{\tilde{p}\ \text{matches the schema }\mathcal{F}\}, \label{eq:cfmt}
\end{align}

With a weight $\alpha=0.9$, the reward is defined as
\begin{equation}
R_{\mathrm{eval}} = \alpha\, c_{\mathrm{all}} + (1-\alpha)\, c_{\mathrm{fmt}}.
\label{eq:reval}
\end{equation}

\subsection{Results and analysis}
\begin{table}[!t]
\centering
\small
\setlength{\tabcolsep}{3.0pt}
\renewcommand{\arraystretch}{0.90}
\resizebox{0.50\textwidth}{!}{%
\begin{tabular}{lccc}
\toprule
\textbf{Model} & \textbf{Overall} & \textbf{Value} & \textbf{Unit} \\
\midrule
\rowcolor{gray!40}
\multicolumn{4}{c}{\textbf{Qwen2.5-VL-7B +}} \\
\addlinespace[1pt]
No RFT (real) & 14.6 & 15.0 & 93.4 \\
GRPO (real) & 19.7 (+34.9\%) & 20.4 (+36.0\%) & 92.3 (-1.2\%) \\
No RFT (synth) & 10.9 & 11.5 & 88.5 \\
GRPO (synth) & 35.2 (+222.9\%) & 35.6 (+209.6\%) & 96.7 (+9.3\%) \\
\addlinespace[2pt]
\rowcolor{gray!40}
\multicolumn{4}{c}{\textbf{Qwen2.5-VL-3B +}} \\
\addlinespace[1pt]
No RFT (real) & 10.5 & 10.8 & 89.3 \\
GRPO (real) & 12.7 (+21.0\%) & 13.8 (+27.8\%) & 89.0 (-0.3\%) \\
No RFT (synth) & 8.4 & 9.1 & 89.9 \\
GRPO (synth) & 31.5 (+275.0\%) & 32.4 (+256.0\%) & 95.7 (+6.5\%) \\
\bottomrule
\end{tabular}
}
\caption{Results of \textsc{Qwen2.5-VL} series with GRPO on real-world and synthetic subsets.}
\label{tab:qwen25vl_grpo}
\end{table}

The evaluation results of \textsc{Qwen2.5-VL} series models with RFT are shown in Table \ref{tab:qwen25vl_grpo}. We get significant performance boost on the in-domain synthetic image test set where the overall accuracy increased by more than threefold (e.g., Qwen2.5-VL-3B: 8.4\% to 31.5\%). Moreover, the model exhibited enhanced generalization to out-of-distribution (OOD) real-world images, with accuracy rising notably (e.g., Qwen2.5-VL-7B: 14.6\% to 19.7\%).

\begin{figure}[!t]
    \centering
    \begin{minipage}[t]{0.48\linewidth}
        \centering
        \includegraphics[width=\linewidth]{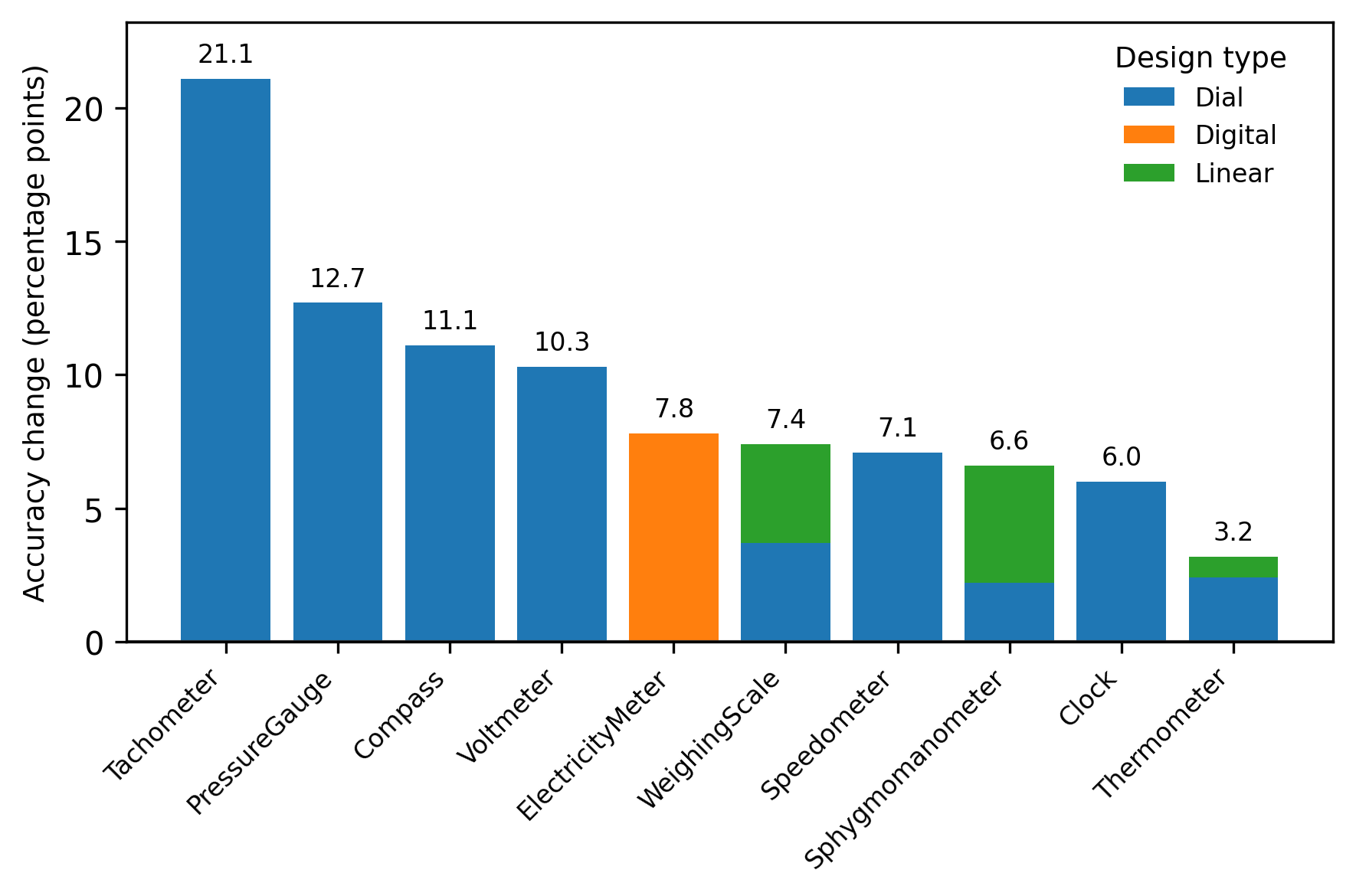}
        \captionof{figure}{Top-10 instrument categories ($\geq$ 20 samples) with the largest accuracy gains after RFT on \textsc{Qwen2.5-VL-7B}.}
        \label{fig:grpo_diff}
    \end{minipage}
    \hfill
    \begin{minipage}[t]{0.48\linewidth}
        \centering
        \includegraphics[width=\linewidth]{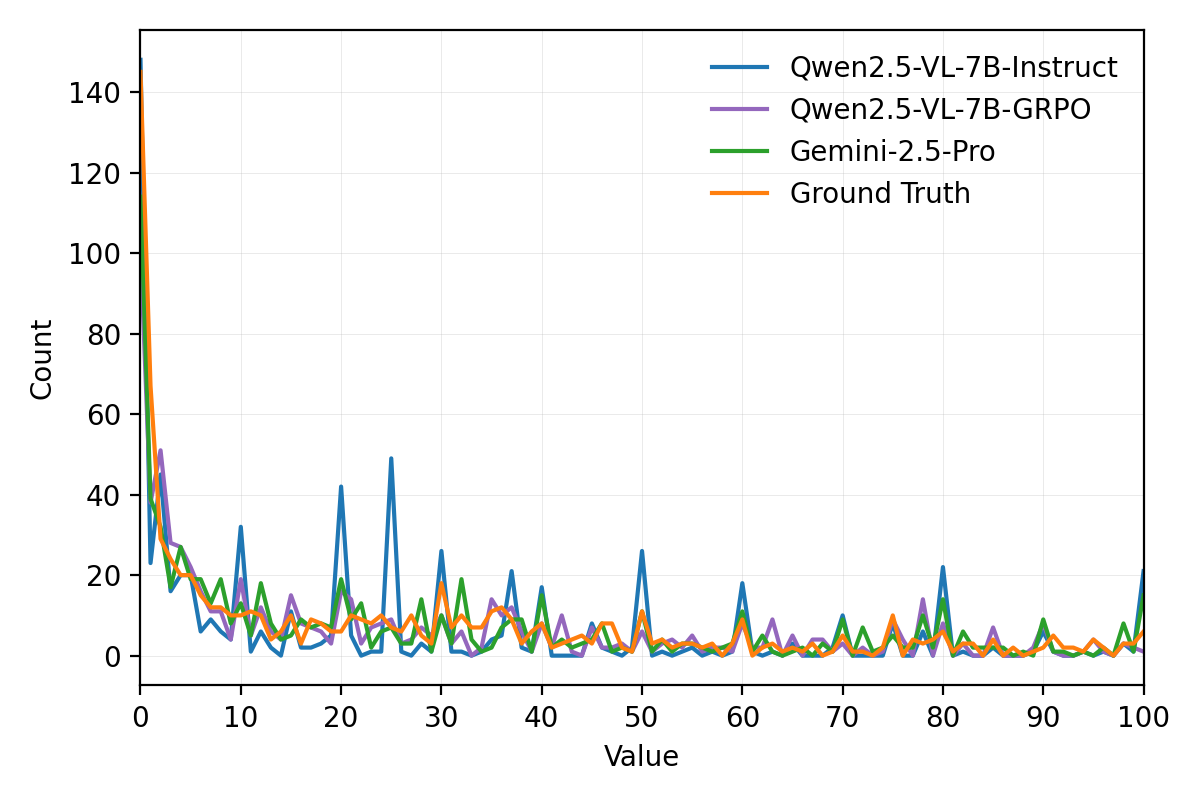}
        \captionof{figure}{Numerical output distributions in the [0, 100] range.}
        \label{fig:dist_change}
    \end{minipage}
\end{figure}

We further analyze the per-instrument accuracy changes on \textsc{Qwen2.5-VL-7B}. As shown in Figure~\ref{fig:grpo_diff}, The top-10 gains occur primarily for simpler instruments, such as single-needle dials (e.g., tachometers), digital readouts (e.g., electricity meters) and linear gauges with sparse ticks. This pattern suggests that RFT on synthetic data is most effective for instruments with low layout complexity, whereas composite reading design remain challenging.

As shown in Figure~\ref{fig:dist_change}, we compare numerical distributions between model predictions and ground truth. The ground-truth readings are relatively smooth, with no pronounced peaks. In contrast, \textsc{Qwen2.5-VL-7B} exhibits clear spikes at round multiples of ten (e.g., 10, 20), indicating a strong language-model prior that favors such values over the visual evidence. After RFT, these peaks are reduced and the distribution becomes smoother and closer to that of a stronger model (e.g., Gemini-2.5-Pro), indicating that RFT helps mitigate this prior bias.

In general, these results show potential from more data curation for VLM training, but also leaving a question on whether we should instead pursuit better model architectures and visual encoding schemes that would make a future VLM genuinely reasoning from detailed visual cues and generalizing over unseen types of instruments.

\section{Related work}

\paragraph{VLMs \& VLM Benchmarks}
Vision–Language Models (VLMs) have made rapid progress in recent years. Early systems such as LLaVA \citep{liu2023llava} and InstructBLIP \citep{dai2023instructblip} pioneered vision instruction tuning, while families like Qwen-VL \citep{Qwen-VL}, InternVL \citep{chen2024internvl}, and GPT-4o \citep{openai_gpt4o_system_card_2024} demonstrated strong general multimodal understanding. More recently, models augmented with reinforcement learning and verifiable rewards (e.g., OpenAI o3 \citep{openai_o3_announce_2025}, Gemini 2.5 Pro \citep{gemini25_2025}, Claude Opus 4 \citep{claude4_system_card}, Qwen3-VL \citep{Qwen2.5-VL}) exhibit improved stepwise reasoning and planning. To assess these capabilities, a broad suite of benchmarks has emerged. General-purpose evaluations (e.g., MMBench \citep{liu2024mmbench}, MM-Vet \citep{yu2024mm}, Seed-Bench \citep{li2023seed,li2023seed2}) target holistic multimodal competence; knowledge-intensive suites (e.g., MMMU \citep{yue2023mmmu, yue2025mmmu-pro}, ScienceQA \citep{lu2022learn}) emphasize academic problem solving; math-centric sets (e.g., MathVision \citep{wang2024measuring}, MathVerse \citep{zhang2024mathverse}) probe visual mathematical reasoning; and perception-focused tests (e.g., CV-Bench \citep{tong2024cambrian1}, BLIND \citep{vlms2024blind}) stress fine-grained visual understanding. More specialized studies on fine-grained reading report persistent weaknesses: SalBench \citep{dahou2025salbench} highlights difficulties with low-level perceptual cues, while BlindTest \citep{vlms2024blind}, SRBench \citep{stogiannidis2025mind}, and VisOnlyQA~\citep{kamoi2025visonlyqa} expose brittle shape, geometry, and spatial reasoning. Despite this progress, relatively less attention has been paid to \emph{instrument reading}, which requires precise localized visual perception coupled with light numerical computation (e.g., inferring tick intervals, decimal placement, and unit normalization).

\paragraph{Measuring Instruments Reading}
Reading measuring instruments is challenging because it integrates fine-grained visual perception, text reading, and visuospatial reasoning. Numerous computer vision methods target specific families of tools such as rulers \citep{pan2025reading}, clocks \citep{yang2022s,saxena2025lost}, water meters \citep{van2025water}, pressure gauges \citep{reitsma2024under}, and other analog dials \citep{howells2021real,salomon2022image,shu2023read,leon2024learning}. Typical pipelines combine detection/segmentation of scales and pointers, geometric rectification, and OCR or tick-interval estimation to map visuals to numeric values and units. However, these approaches are narrowly tailored on moderate-scale training/validation data henceforth generalize poorly across device types, design variations, viewpoints, glare/occlusion, and unit ambiguity. More recently, VLMs have been applied to instrument reading: GPT-4o \citep{openai_gpt4o_system_card_2024} reports preliminary ability on industrial gauges, and CAD2DMD-SET \citep{valente2025cad2dmd} evaluates several VLMs on digital measurement devices. Yet current evaluations remain fragmented: they cover limited device diversity, emphasize categorical correctness over calibrated numeric error, and seldom assess unit normalization, tolerance bands, or robustness stressors. %

\section{Conclusions and Discussion}
We introduced \textsc{MeasureBench}, a comprehensive benchmark with both real-world and synthetic subsets for evaluating vision–language models (VLMs) on instrument reading. Our analyses reveal a persistent limitation of current VLMs: difficulty with fine-grained visual cues and precise visual–numeric correspondences, leading to errors in value estimation and unit normalization. The proposed synthetic data generation pipeline serves both as a source of controlled benchmarks and as an effective means of training data augmentation. We also explored reinforcement finetuning with GRPO. Preliminary results suggest that even small amounts of targeted synthetic data can yield measurable gains that transfer to real-world settings, but only to a moderate extent. We hope this work could help future VLM development with more comprehensive training data curation or better visual representation modeling to enable stronger capabilities in fine-grained understanding, geometric alignment, and spatial reasoning.

\section*{Authors}

Fenfen Lin$^*$, Yesheng Liu$^*$, Haiyu Xu$^*$, Chen Yue$^*$, Zheqi He\renewcommand{\thefootnote}{$\dagger$}\footnote{Project lead; $^*$~equally contributed to this work. correspondance to: zqhe at baai.ac.cn}, Mingxuan Zhao, Miguel Hu Chen, Jiakang Liu, JG Yao, Xi Yang

\bibliography{main}
\bibliographystyle{flageval_baai}

\appendix
\section{Additional Details}%

\subsection{Training details}
\label{app:training}

We employ reinforcement finetuning (RFT) on the synthesis datasets with 3900 images by reinforcement learning framework verl. Following Deepseek-R1, we employ GRPO with a format reward function to optimize the model to output the thinking process within ``\texttt{<think>...</think>}".
Training is performed on 8×H100 GPUs for 15 epochs with a global batch size of 128 and a learning rate of $1\times10^{-6}$, and a rollout number of 8.

\subsection{Soft-margin Reward}
We also conduct experiments with a soft-margin reward that grants partial credit to predictions near the target interval. For the numeric component of the answer, define the distance to the interval as
\begin{equation}
d(\hat{y}, I) =
\begin{cases}
0, & \hat{y} \in [l, r],\\[2pt]
\min(|\hat{y} - l|,\, |\hat{y} - r|), & \text{otherwise},
\end{cases}
\end{equation}
and the margin
\begin{equation}
m(I) =
\begin{cases}
r - l, & r > l,\\[2pt]
0.05\,l, & \text{otherwise},
\end{cases}
\end{equation}

with a small constant $\varepsilon>0$.
We define a linearly decaying partial-credit score
\begin{equation}
s_{\mathrm{sm}}(\hat{y}; I) \;=\; \tfrac{1}{2}\,\max\!\Bigl\{0,\; 1 - \tfrac{d(\hat{y}, I)}{\,m(I)+\varepsilon\,}\Bigr\}.
\end{equation}
The soft-margin reward replaces the value term by taking the better of exact correctness and soft credit, while keeping the formatting term unchanged:
\begin{equation}
R_{\mathrm{soft}} \;=\; \alpha\, \max\!\bigl\{\,c_{\mathrm{all}},\; s_{\mathrm{sm}}(\hat{y}; I)\,\bigr\}
\;+\; (1-\alpha)\, c_{\mathrm{fmt}}.
\end{equation}

This keeps rewards consistent with evaluation when the answer is exact, while giving informative feedback to near-miss predictions.

\begin{table*}[ht]
\centering
\small
\begin{tabular}{l|ccc}
\toprule
\textbf{Model/Dataset} & \textbf{Overall} & \textbf{Value} & \textbf{Unit} \\
\midrule
Qwen2.5-VL-7B (Real-world) & 14.6 & 15.0 & 93.4 \\
\rowcolor{gray!10}
Qwen2.5-VL-7B+GRPO (Real-world) & 19.7 (+34.9\%) & 20.4 (+36.0\%) & 92.3 (-1.2\%) \\
Qwen2.5-VL-7B+GRPO-soft (Real-world) & 19.6 (+34.2\%) & 20.4 (+36.0\%) & 91.9 (-1.6\%) \\
\rowcolor{gray!10}
Qwen2.5-VL-7B (Synthetic) & 10.9 & 11.5 & 88.5 \\
Qwen2.5-VL-7B+GRPO (Synthetic) & 35.2 (+222.9\%) & 35.6 (+209.6\%) & 96.7 (+9.3\%) \\
\rowcolor{gray!10}
Qwen2.5-VL-7B+GRPO-soft (Synthetic) & 35.3 (+223.9\%) & 35.6 (+209.6\%) & 98.0 (+10.7\%) \\
\bottomrule
\end{tabular}
\caption{Results of Qwen2.5-VL-7B variants with GRPO on real-world and synthetic subsets.}
\label{tab:qwen25vl_grpo_soft}
\end{table*}

As shown in Table~\ref{tab:qwen25vl_grpo_soft}, the soft-margin reward yields comparable performance to the original hard-margin reward after RFT on both real-world and synthetic subsets. This suggests that the original reward design is already effective, while the soft-margin variant provides an alternative that may be more suitable in scenarios where near-miss predictions are common.

\section{Examples of synthesized images}

\begin{figure*}[!ht]
  \centering
  \includegraphics[width=1.0\textwidth]{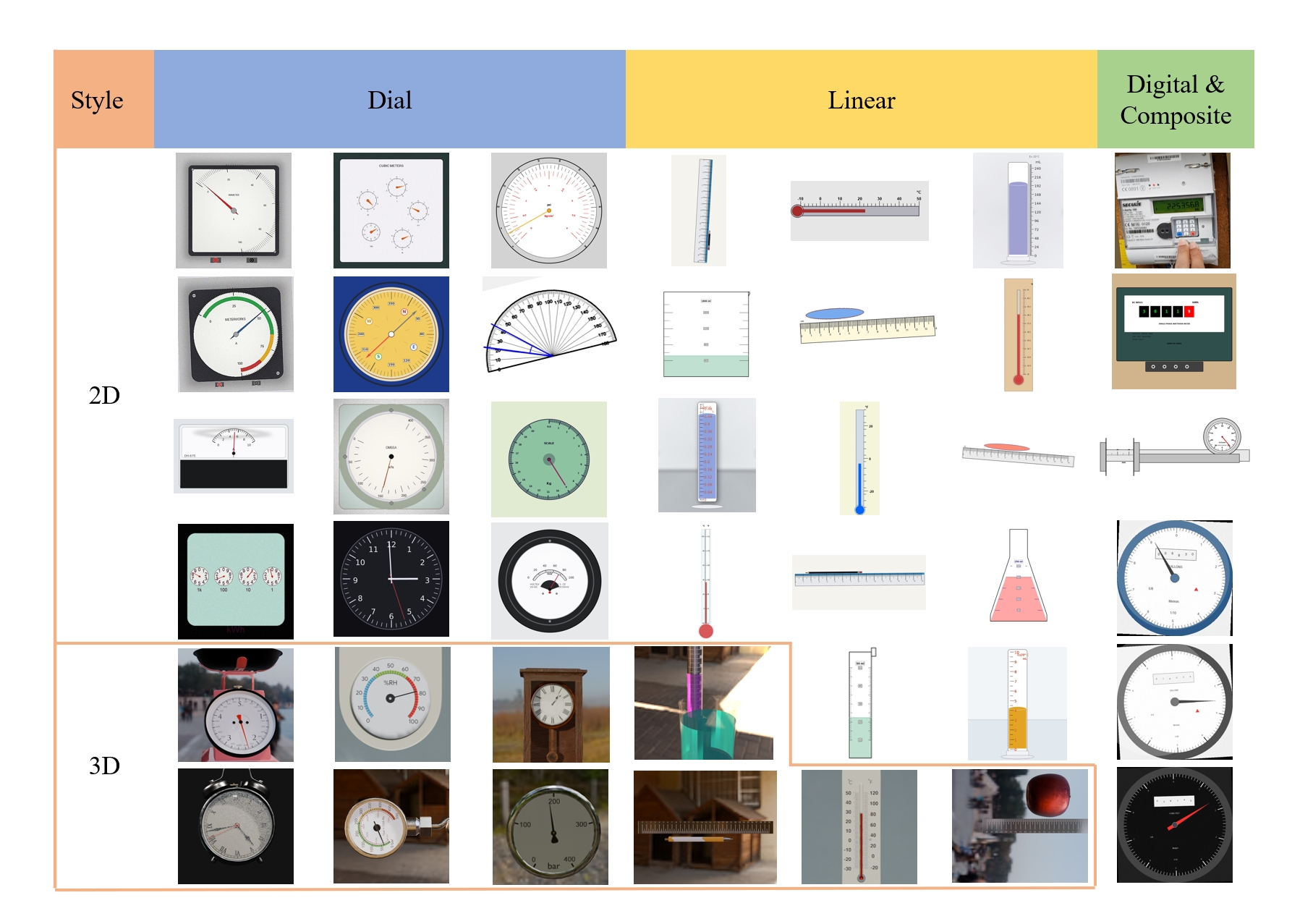}
  \caption{Additional examples of synthetic measuring instruments generated by our pipeline. }
  \label{fig:appendix_instruments}
\end{figure*}

Here we provide additional examples of synthesized images of measuring instruments generated by our framework. Each generator in our framework is expected to render an image of an instrument with similar appearance along with random readout. In Figure \ref{fig:appendix_instruments}, 2D images are rendered by offline-only libraries like Pillow, NumPy and Matplotlib, 3D images are rendered by Blender which is more realistic.

\section{3D modeling with Blender}
\label{sec:blender-details}

To construct a large collection of measurement-related 3D assets, we use  \href{https://www.blender.org/download/releases/4-2/}{\textbf{Blender (v4.2)}} in combination with publicly available online repositories. The procedure was as follows.
\begin{figure*}[!ht]
    \centering
    \includegraphics[width=1.0\textwidth]{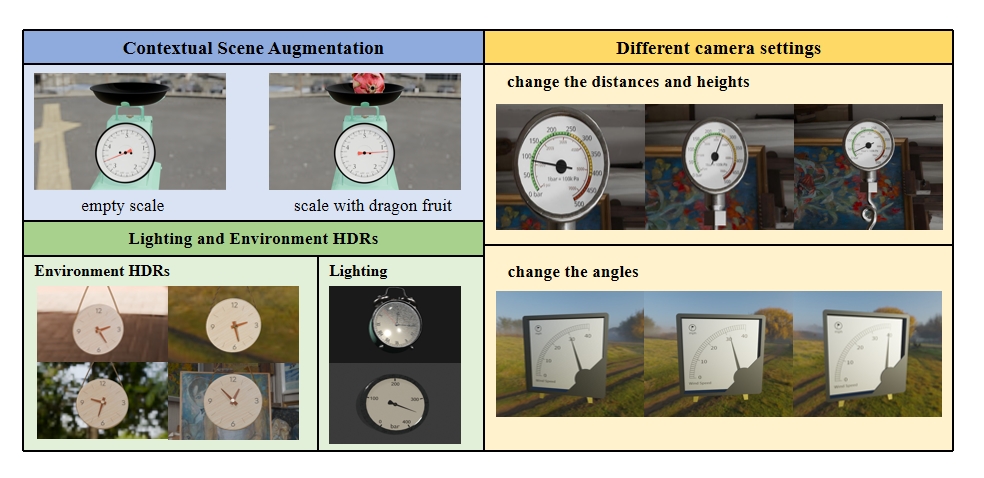}
    \caption{Examples of augmentation strategies applied during 3D model acquisition and preparation with Blender (v4.2).}
    \label{fig:Blender_illustrations}
\end{figure*}
\subsection{Asset retrieval}
We integrate the \href{https://www.blenderkit.com/}{\textbf{BlenderKit}} plugin into Blender to access free 3D assets, including models, HDRs, and materials. For categories underrepresented in BlenderKit (e.g., cylinder, hygrometer), we also retrieved models from \textbf{Sketchfab}. The queries included \textit{watches, clocks, scales} and \textit{rulers, thermometers}, covering both \textbf{pointer-based} and \textbf{linear-scale instruments}.

\subsection{Model normalization}
\begin{enumerate}
    \item \textbf{Pointer-based models} (e.g., clocks, scales): In many assets, the pointer was not initially aligned with the zero position. We manually rotated the pointer to  zero and reset its transformations (rotation along the x, y, z axes set to 0).
    \item \textbf{Linear-scale models} (e.g., thermometers): For these, we determined the minimum-maximum mapping on the scale and adjusted the geometry proportionally so that the linear transformations of pointer correctly represented measurement values.
\end{enumerate}

\subsection{Contextual scene augmentation}
Some models only represented the measurement instrument itself, which led to unrealistic renderings when the pointer indicated a nonzero value. To improve semantic consistency, we augmented scenes with additional objects:
\begin{itemize}
    \item \textbf{Scales}: To avoid showing a dial reading \textit{1 kg} with an empty plate, we placed an additional object (e.g., a fruit model, such as dragon fruit) on the weighing surface.
    \item \textbf{Rulers}: Since rulers measure relative length, we included a reference object (a pen). The pen was rescaled and positioned alongside the ruler, allowing queries such as "How long is the pen?" to be grounded in the rendered image.
\end{itemize}
These contextual additions ensured that pointer readings were visually consistent with the surrounding scene, enhancing dataset realism, and reducing ambiguity for vision-language evaluation.

\subsection{Pointer rotation control}
Pointer manipulation was automated with Blender's Python API.
\begin{itemize}
    \item \textbf{Clocks}: For clocks, rotation angles were computed directly from the target hour, minute, and (optionally) second values:
    \begin{lstlisting}[style=codeblock]
second_angle = math.radians(target_second * 6)
minute_angle = math.radians(target_minute * 6 + target_second * 0.1)
hour_angle   = math.radians((target_hour %
    \end{lstlisting}
    The axis of rotation varied across different models (i.e. whether Oxy, Oxz, or Oyz). For example, a clock's hour hand can be controlled with: \texttt{hour\_hand.rotation\_euler = (0, 0, -hour\_angle)}. However, depending on the model, the rotation angle might be applied to the first or second component of the Euler tuple rather than to the third.
    \item \textbf{Other dials} (e.g., hygrometers): For these, the degree of pointer rotation depends on the specific model geometry.
    We first check for the maximum rotation angle ($max.rot.deg$) that corresponded to the maximum scale value, and set pointer positions linearly:
    \begin{lstlisting}[style=codeblock]
max_rot = math.radians(max_rot_deg)
rot_z = min_rot + (humidity-min_humidity) / (max_humidity-min_humidity)
            * (max_rot-min_rot)
    \end{lstlisting}
    This approach is generalized to other instruments with linear or semi-linear dial mappings.
\end{itemize}
If the geometry of the model used a nonstandard orientation, we rotated the entire object to align it with the desired axis.

\subsection{Camera alignment}
Since the dial panels of many models were not centered at the origin, we applied offsets to position the camera such that it directly faces the dial. Camera distance and angle were tuned empirically to maximize legibility of the dial face and pointer. For small-scale instruments, shorter distances and narrower angle ranges provided clearer renderings, whereas larger instruments benefited from wider perspectives.

\subsection{Lighting and environment HDRs}
To ensure consistent illumination across renderings, we used two strategies depending on the dataset requirements:
\begin{itemize}
    \item \textbf{HDR environment maps}: For most models, we initialized scenes with background environment maps (.exr files), either using Blender's built-in HDRIs or downloading additional ones via BlenderKit. These provided realistic lighting and surface reflections. HDRs were first manually configured and later automated using Python.
    \item \textbf{Direct light sources}: For cases where a clean background was preferred, we disabled HDRs and instead added light objects from different positions (e.g., point lights or area lights). This clearly illuminated the dial while leaving the background neutral.
\end{itemize}

\subsection{Rendering execution}
Scripts were executed either directly within Blender's Scripting panel or externally via Python (importing the bpy module) in an IDE such as Visual Studio Code. This flexibility enabled large-scale automated rendering of models across different instrument categories.

Figure \ref{fig:Blender_illustrations} provides illustrative examples of the augmentation strategies described above.

\section{Extended Analysis}
We provide further analysis of model behavior on specific challenging cases, as well as statistical distributions of numerical outputs.

\subsection{Reading of complex measuring instruments}
\begin{figure*}[!ht]
    \centering
    \includegraphics[width=0.9\linewidth]{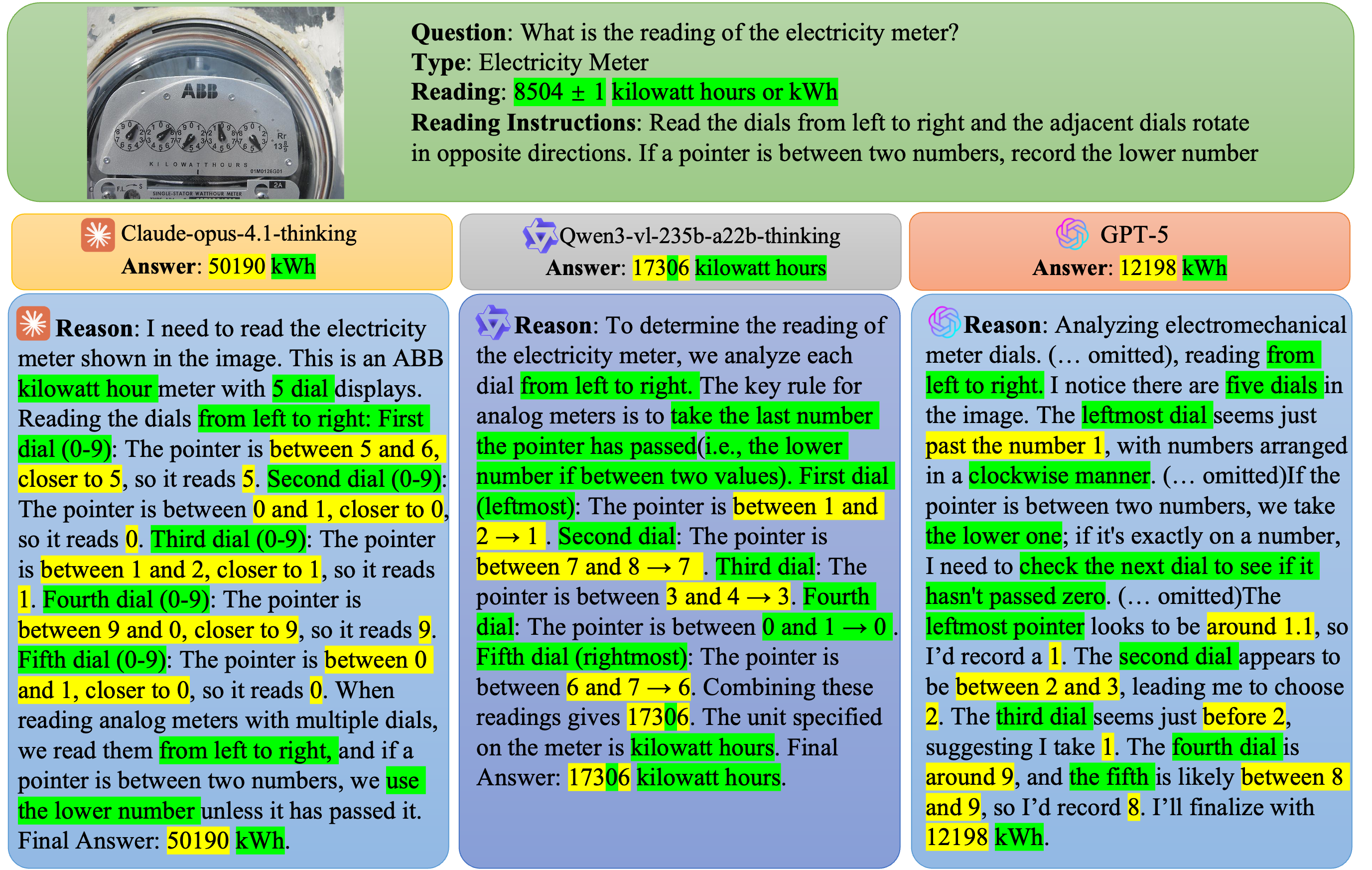}
    \caption{Results comparison on an electricity meter}
    \label{fig:electricity_meter_example}
\end{figure*}

Complex measuring instruments with composite readout designs or multiple dials and pointers remain highly challenging for current VLMs: an error at any step of fine-grained visual perception or reading interpretation will propagate to an incorrect final result.
Figure~\ref{fig:electricity_meter_example} shows a multi-dial electricity meter with five dials, each with a pointer. To read the meter, one should note the position of each pointer and record the numbers from \textbf{left to right}, remembering that adjacent dials rotate in opposite directions. If a pointer is between two numbers, the \textbf{lower} number is recorded, unless it is between 9 and 0, in which case 9 is recorded. We present the results of three models, although all of them correctly describe the reading procedure, they still struggle to localize the pointer positions, misreading almost all pointers and thus producing incorrect final readings.

\subsection{Example of correct ``guessing''}
\begin{figure*}[!ht]
    \centering
    \includegraphics[width=0.85\linewidth]{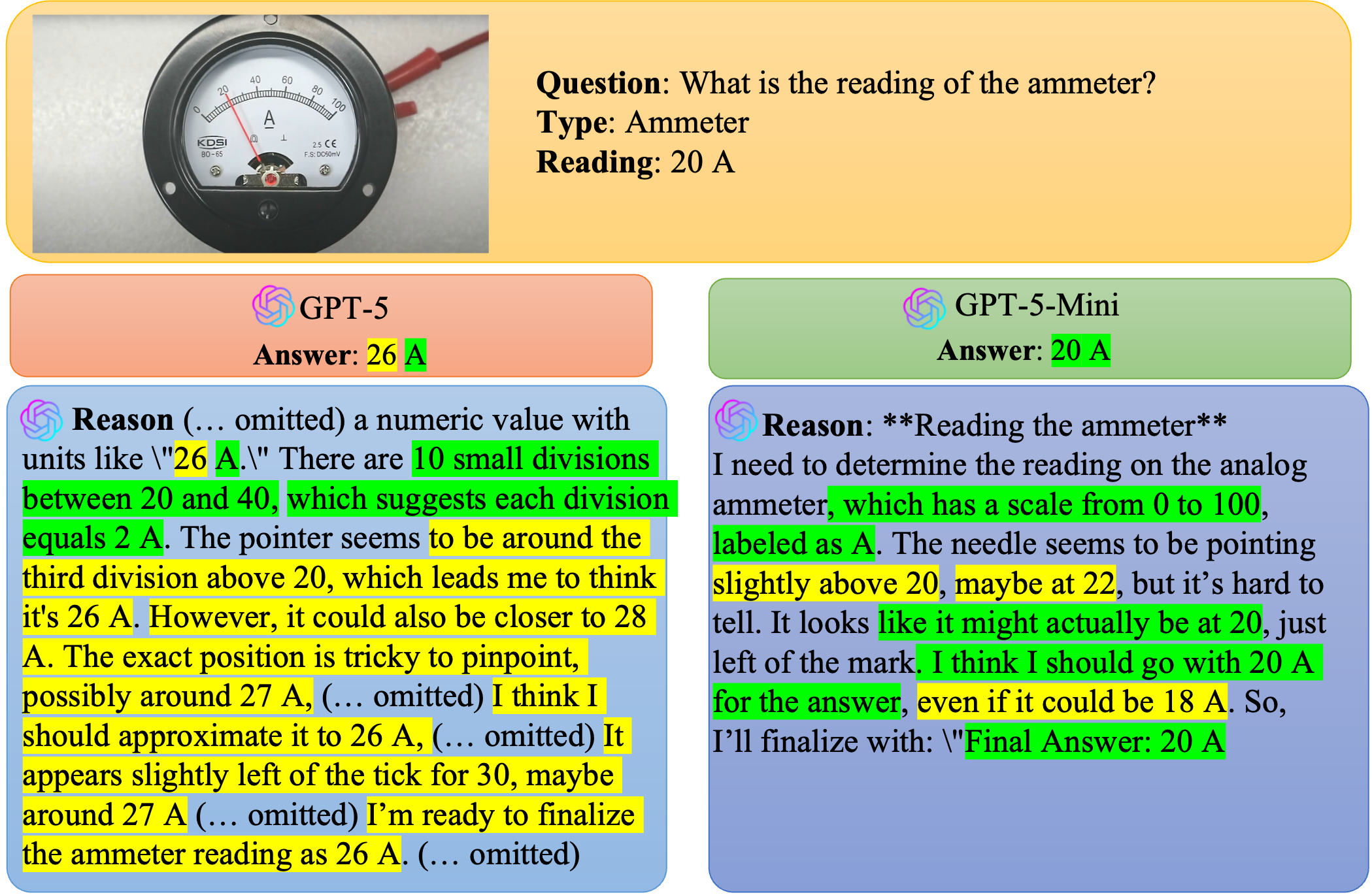}
    \caption{An example where GPT-5 answers incorrectly while GPT-5-Mini guesses the correct reading.}
    \label{fig:/gpt-5-mini-guess}
\end{figure*}
We observed that some correct answers may arise from guessing rather than accurate visual understanding. Figure~\ref{fig:/gpt-5-mini-guess} shows an ammeter whose pointer indicates a reading of 20~A. Both GPT-5 and GPT-5-Mini provide detailed reasoning steps before giving their final answers. GPT-5 incorrectly estimates the reading as 26~A, whereas GPT-5-Mini is clearly uncertain and essentially guesses 20~A, which happens to be correct.

\section{The ``10:10'' phenomenon}
\label{app:1010}
\begin{table}[!ht]
\centering
\begin{tabular}{lcc}
\toprule
Model & Real-world & Synthetic \\
\midrule
Qwen2.5-VL-72B-Instruct & 72.88\% & 50.74\% \\
GPT-5-Mini & 29.66\% & 7.78\% \\
Claude-Sonnet-4 & 26.27\% & 16.30\% \\
Qwen2.5-VL-7B-Instruct & 23.73\% & 15.56\% \\
Qwen2.5-VL-32B-Instruct & 21.19\% & 8.89\% \\
GPT-5 & 20.34\% & 6.30\% \\
Mistral-medium-3.1 & 16.95\% & 4.07\% \\
InternVL3.5-38B-thinking & 16.10\% & 12.96\% \\
Claude-Opus-4.1 & 13.56\% & 9.63\% \\
InternVL3.5-8B-thinking & 12.71\% & 7.04\% \\
Claude-Opus-4.1-thinking & 12.71\% & 10.37\% \\
Qwen3-VL-235b-instruct & 12.71\% & 12.96\% \\
InternVL3.5-38B & 11.86\% & 10.00\% \\
Qwen2.5-VL-3B-Instruct & 11.86\% & 3.70\% \\
Gemini-2.5-Pro & 11.86\% & 3.33\% \\
Qwen3-VL-8B & 7.63\% & 9.63\% \\
Gemini-2.5-Flash & 4.24\% & 1.11\% \\
InternVL3.5-8B & 3.39\% & 6.30\% \\
\textbf{Qwen2.5-VL-7B-GRPO} & 3.39\% & 1.11\% \\
Grok-4 & 3.39\% & 4.81\% \\
Gemini-2.5-Flash-thinking & 2.54\% & 1.48\% \\
LLaMA-4-maverick & 0.85\% & 1.11\% \\
LLaMA-4-scout & 0.00\% & 1.85\% \\
\bottomrule
\end{tabular}
\caption{Proportion of "10:10" responses on clock images in MeasureBench.}
\label{tab:clock_1010_stats}
\end{table}

We found an interesting phenomenon that many models tend to answer ``10:10'' when reading clock times, regardless of the actual time shown in the image.  During the real-world data collection, we deliberately avoid including images with ``10:10'', and in the synthesis process,clock times are uniformly sampled. As a result, there are few ground-truth of ``10:10'' in our benchmark.

However, as shown in Table~\ref{tab:clock_1010_stats}, we calculate the proportion of string ``10:10'' in the models's answers on clock images. It is surprising that the powerful open-source model Qwen2.5-VL-72B-Instruct outputs ``10:10'' for 72.88\% of real-world clock images and 50.74\% of synthetic clock images. The frontier commercial model GPT-5 also predicts ``10:10'' in more than 20\% of its answers on real-world clock images. This bias likely stems from training data, where clocks are frequently depicted at ``10:10'' for aesthetic reasons in advertisements and product listings. To further verify this, we examine the answers of Qwen2.5-VL-7B with RFT training on our synthetic dataset, where clock times follow a uniform distribution.  The RFT-trained model predicts ``10:10'' on only 3.39\% of real-world clock images, whereas the original Qwen2.5-VL-7B does so on 23.73\% of images, indicating that training with a more uniform distribution can effectively mitigate this bias.

\section{Spikes distributions at integers}
\begin{figure}[!htp]
    \centering
    \begin{minipage}[t]{0.48\linewidth}
        \centering
        \includegraphics[width=\linewidth]{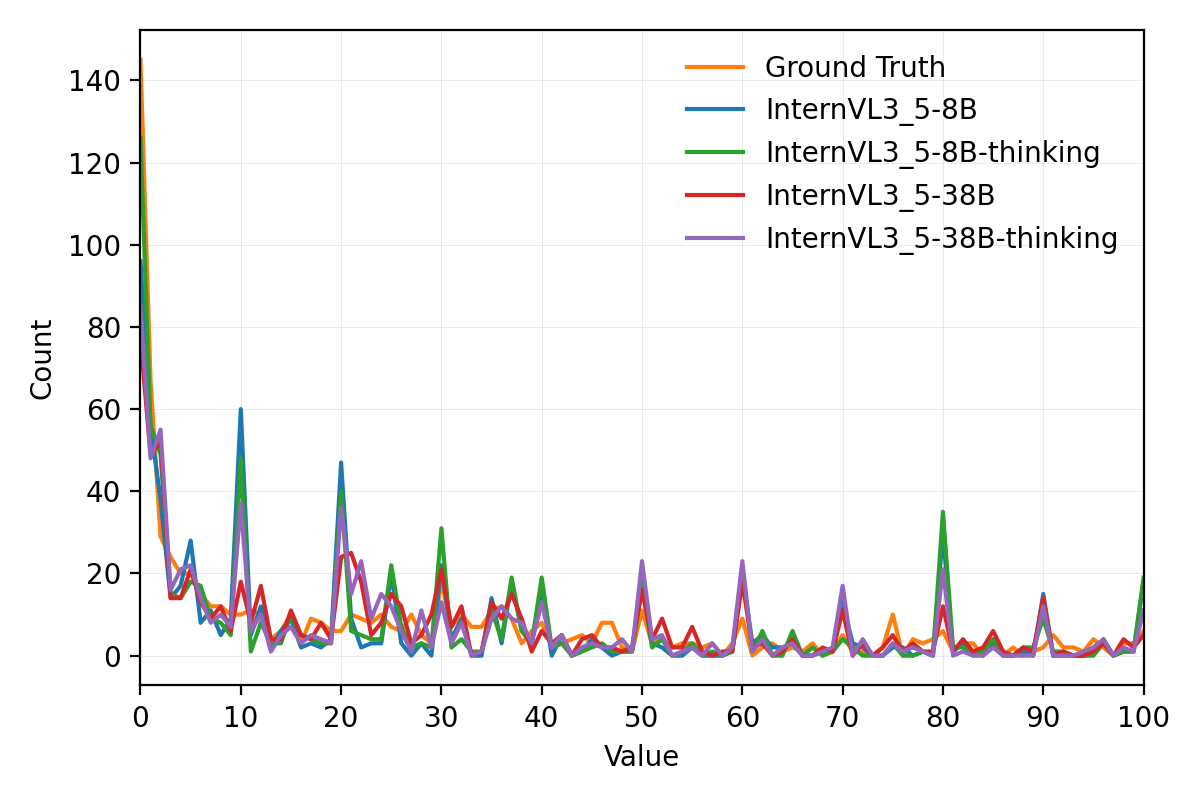}
        \captionof{figure}{The numeric spikes distribution of InternVL3.5 series on real-world subset.}
        \label{fig:dis_internvl}
    \end{minipage}
    \hfill
    \begin{minipage}[t]{0.48\linewidth}
        \centering
        \includegraphics[width=\linewidth]{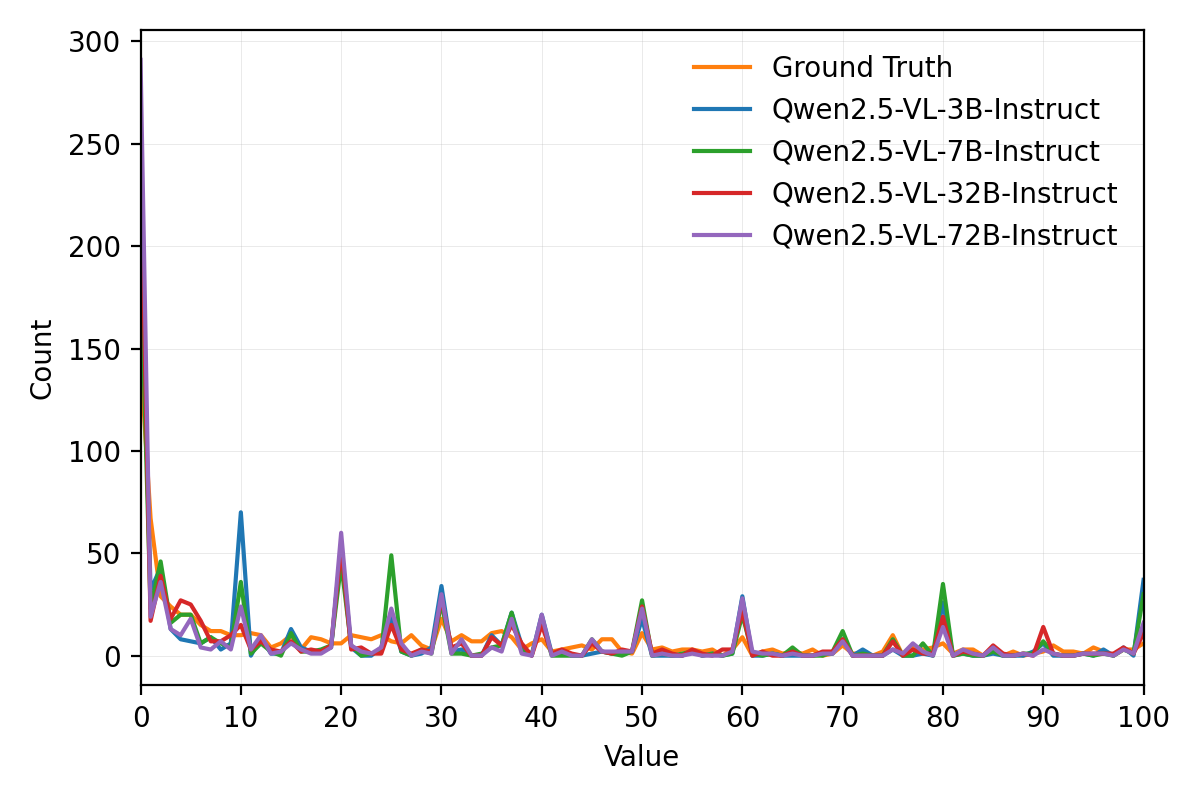}
        \captionof{figure}{The numeric spikes distribution of Qwen2.5-VL series on real-world subset.}
        \label{fig:dis_qwen}
    \end{minipage}
\end{figure}
We further analyze the distribution of numeric outputs from different models. Figure~\ref{fig:dis_internvl} and Figure~\ref{fig:dis_qwen} show the spikes at integer values for InternVL3.5 series and Qwen2.5-VL series respectively. We can observe that both models exhibit significant spikes at multiples of ten, the same model series have similar distribution patterns, and thinking mode can not mitigate these spikes. This may be attributed to the models' training data, where round numbers are more frequently represented.

\end{document}